\documentclass{article}
\usepackage{iclr2027_conference,times}

\usepackage{amsmath,amsfonts,bm}

\def\eqref#1{equation~\ref{#1}}

\def\1{\bm{1}}

\DeclareMathAlphabet{\mathsfit}{\encodingdefault}{\sfdefault}{m}{sl}
\SetMathAlphabet{\mathsfit}{bold}{\encodingdefault}{\sfdefault}{bx}{n}

\DeclareMathOperator*{\argmax}{arg\,max}

\usepackage{hyperref}
\hypersetup{
  hidelinks,
  pdftitle={Contrastive On-Policy Distillation},
  pdfauthor={Jiacheng Ruan, Jun Tang, Wenzhen Yuan, Ting Liu, Shuai Bai, Dayiheng Liu, Zhibo Yang, Yuzhuo Fu}
}
\usepackage{url}
\usepackage{graphicx}
\usepackage{booktabs}
\usepackage{array}
\usepackage{multirow}
\usepackage{amsmath}
\usepackage{amssymb}
\usepackage{float}
\usepackage{wrapfig}
\usepackage{caption}
\usepackage{algorithm}
\usepackage{algpseudocode}
\usepackage{tcolorbox}
\tcbuselibrary{breakable}
\usepackage{needspace}

\title{Contrastive On-Policy Distillation}

\author{
\normalfont\textbf{Jiacheng Ruan}\textsuperscript{1,2}\thanks{This work was done during an internship with the Qwen Team at Alibaba Group.},\quad
\textbf{Jun Tang}\textsuperscript{2},\quad
\textbf{Wenzhen Yuan}\textsuperscript{1},\quad
\textbf{Ting Liu}\textsuperscript{1},\\
\normalfont\textbf{Shuai Bai}\textsuperscript{2},\quad
\textbf{Dayiheng Liu}\textsuperscript{2},\quad
\textbf{Zhibo Yang}\textsuperscript{2}\thanks{Project leader.},\quad
\textbf{Yuzhuo Fu}\textsuperscript{1}\thanks{Corresponding author.}\\
\normalfont\textsuperscript{1}Shanghai Jiao Tong University,\quad
\textsuperscript{2}Alibaba Group\\
\normalfont\texttt{jackchenruan@sjtu.edu.cn}
}

\newcommand{\method}{COPD}
\newcommand{\selfmethod}{COPSD}
\newcommand{\lt}{\mathrm{LT}}
\newcommand{\htmode}{\mathrm{HT}}
\newcommand{\accperk}{Acc@1K}
\newcommand{\ms}[2]{$#1_{\pm#2}$}
\newcommand{\msb}[2]{$\mathbf{#1}_{\pm#2}$}

\iclrfinalcopy

\begin{document}

\maketitle
\lhead{Contrastive On-Policy Distillation}

\begin{abstract}

On-policy distillation (OPD) trains a student model on trajectories sampled from its own policy, providing dense token-level supervision by minimizing the divergence between the teacher's and student's output distributions at each token. Although existing OPD approaches effectively distill strong reasoning capabilities into student models, they inherently inherit uncurated chain-of-thought traces, exacerbating overthinking and reasoning redundancy. To address this limitation, we introduce \method{}, a contrastive OPD framework. Specifically, for each token generated by the student, a frozen teacher evaluates the current student state under two contrasting prompts that induce low and high reasoning effort, respectively. The resulting difference in log-probabilities serves as a token-level advantage signal to guide the policy update. Rather than merely mimicking a single teacher distribution, \method{} guides the student model toward acquiring concise and efficient reasoning strategies. We evaluate \method{} across 9 multimodal benchmarks covering both reasoning and understanding tasks. Empirical results show that \method{} substantially reduces reasoning length without hurting task performance, consistently improving efficiency across different tasks and model scales. In addition, this contrastive paradigm extends naturally to on-policy self-distillation (OPSD), establishing self-contrastive supervisory signals that enable a single model to compress its own reasoning without an external teacher.

\end{abstract}

\section{Introduction}

Large vision-language models (LVLMs) have made substantial progress in multimodal tasks \citep{flamingo2022,blip22023,llava2023,liu2023llava15,chen2023internvl,bai2023qwenvl,wang2024qwen2vl,bai2025qwen25vl,qwen3vl2025,caffagni2024mllmsurvey,jin2024efficientmllmsurvey}. By decomposing complex problems into explicit intermediate steps, these models steadily improve their performance in tasks such as image understanding and vision-language reasoning \citep{wei2023chainofthought,kojima2022zeroshot,zhou2022leasttomost,zhang2023multimodalcot}. Consequently, step-by-step reasoning has become the primary paradigm for complex problem solving \citep{wei2023chainofthought,yao2023tree,huang2023reasoningsurvey}. However, stronger reasoning capabilities often introduce severe redundancy into the reasoning chains. A central challenge is therefore to improve reasoning efficiency without sacrificing performance.

Existing reasoning-compression methods predominantly rely on supervised fine-tuning (SFT) or reinforcement learning with verifiable rewards (RLVR) \citep{munkhbat2025concise,jin2025recut,han2024tokenbudget,li2025selfbudgeter,liu2025dler,li2025leash,dumitru2025conciserl}. However, they either demand scarce, curated concise traces or require ground-truth verifiers, limiting their applicability to unlabeled scenarios. On-policy distillation (OPD) provides a label-free alternative by distilling teacher feedback along student-generated sequences \citep{gkd2024,lu2025onpolicydistillation}. Yet, standard OPD aligns the student with only a single teacher distribution, failing to capture how token compatibility shifts across different reasoning efforts.

To this end, we introduce \method{}, a contrastive on-policy distillation framework. By pairing light- and heavy-thinking (LT/HT) contexts, this approach converts the implicit preferences of the teacher into explicit token-level supervisory signals. Specifically, the student model first generates reasoning trajectories on-policy. A frozen teacher then evaluates this trajectory at the token level under two contrasting prompts specifying low and high reasoning effort, respectively. The difference between the resulting log-likelihoods serves as a token-level advantage signal to guide student optimization. Under this design, the teacher functions not merely as an imitation target for a single distribution, but as a contrastive reference across different reasoning efforts. Consequently, this contrastive signal biases the student toward concise outputs aligned with the light-thinking condition, enabling adaptive reasoning compression without predefined length budgets.

We systematically evaluate \method{} across 9 multimodal benchmarks. Empirical results show that across diverse task types and model scales, \method{} substantially shortens reasoning length while maintaining or improving task accuracy, thereby optimizing the trade-off between performance and inference cost. For instance, with Qwen3-VL-8B as the teacher and Qwen3-VL-2B \citep{qwen3vl2025} as the student, \method{} improves accuracy from 55.2\% to 57.0\% compared to ExOPD \citep{exopd2026}, while cutting response length by 56.3\%. In addition, this contrastive mechanism extends naturally to on-policy self-distillation (OPSD), constructing self-contrastive supervisory signals without an external teacher to compress reasoning lengths effectively.

The main contributions of this paper are summarized as follows: \textbf{1)} We introduce \method{}, a contrastive on-policy distillation framework that derives dense token-level advantage signals from paired teacher evaluations under contrasting light- and heavy-thinking instructions, achieving adaptive reasoning compression with negligible performance loss. \textbf{2)} We extend this contrastive paradigm to on-policy self-distillation (OPSD), enabling models to self-induce concise and efficient reasoning strategies entirely free of external supervision. \textbf{3)} Extensive experiments across 9 multimodal benchmarks demonstrate the consistent effectiveness and robust generalization of \method{} across diverse tasks and model scales.

\section{\method{}: Contrastive On-Policy Distillation}
\label{sec:method}

\subsection{Problem setup and overview}

Let $x$ denote an input prompt, $\pi_\theta$ the student policy, and $\pi_\phi$ a frozen teacher model. On-policy distillation samples a response $y=(y_1,\dots,y_{T_y})$ of length $T_y=|y|$ from the student policy $\pi_{\theta_{\mathrm{old}}}$ and obtains teacher supervision at each student-visited prefix $(x,y_{<t})$ \citep{gkd2024,lu2025onpolicydistillation}. While this mechanism provides dense supervision along student-generated trajectories, its training signal is derived from a single teacher conditional distribution at each visited prefix.

\begin{figure}[t]
\centering
\includegraphics[width=\linewidth]{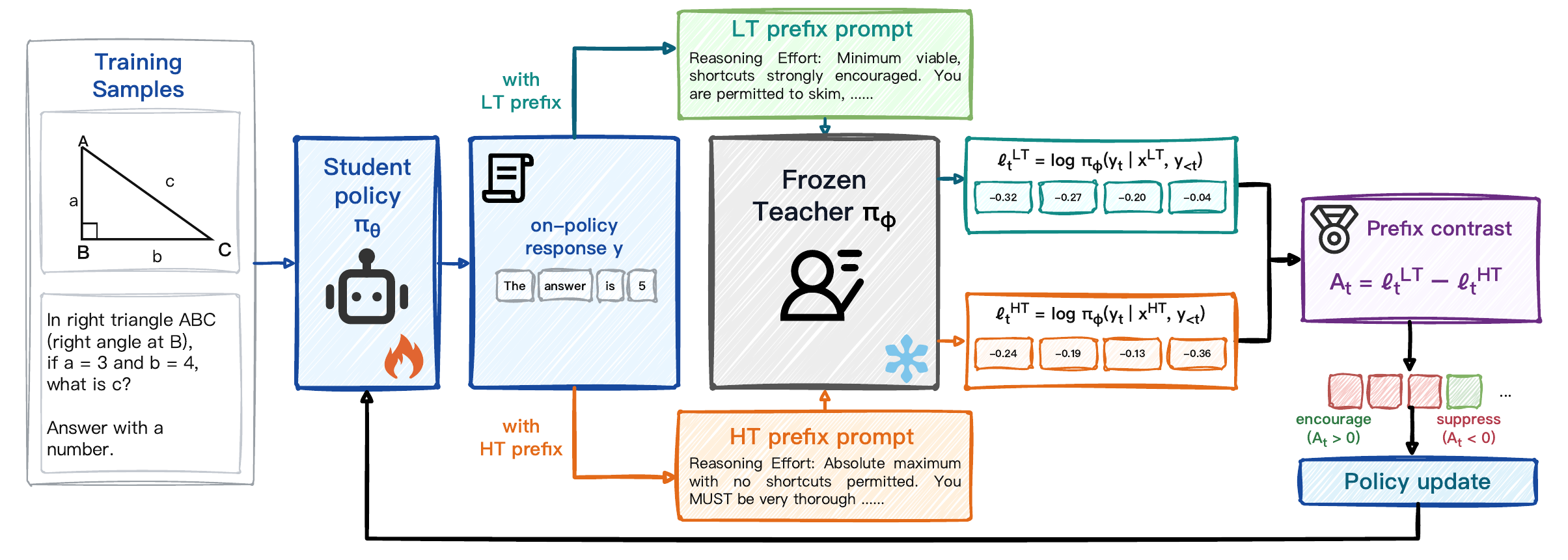}
\caption{\textbf{Overview of the \method{} training pipeline.} The student model first samples a response trajectory given the input. A frozen teacher then evaluates the generated sequence token by token under contrasting light- and heavy-thinking (LT and HT) prompts. The resulting difference in log-probabilities serves as a token-level advantage signal to guide the student policy update.}
\label{fig:method}
\end{figure}

As shown in Figure \ref{fig:method}, \method{} casts teacher scoring as a paired comparison across different reasoning efforts. Specifically, given a student-generated trajectory, the frozen teacher scores each token conditioned on the shared multimodal input and the preceding prefix, differing solely in the light- versus heavy-thinking prompt. The difference between the resulting log-likelihoods constitutes a token-level preference signal. Consequently, \method{} preserves the student-induced state distribution inherent to OPD, while replacing single-context imitation with a dense estimate of whether the current student action aligns more closely with concise or exhaustive reasoning.

\subsection{Paired teacher scoring}

Let $\iota^{\lt}$ and $\iota^{\htmode}$ denote light- and heavy-thinking instructions, respectively. Given input $x$, the student policy first samples a complete response trajectory:
\begin{equation}
    y = (y_1, \dots, y_{T_y}) \sim \pi_{\theta_{\mathrm{old}}}(\cdot \mid x),
\end{equation}
Subsequently, the frozen teacher receives the same input and evaluates the identical trajectory $y$, computing token-level conditional log-probabilities under both light- and heavy-thinking instructions:
\begin{equation}
    \ell_t^m=\log \pi_\phi(y_t\mid x^m,y_{<t}),
    \qquad x^m=[\iota^m;x],\quad m\in \{\lt,\htmode\},\quad t=1,\dots,T_y.
\end{equation}
Intuitively, the same token can exhibit markedly different conditional likelihoods depending on the intended reasoning effort \citep{han2024tokenbudget,shi2024trusting}. Paired scoring builds directly on this observation: through two forward passes of the frozen teacher, it evaluates the identical generation step under contrasting instruction contexts, yielding the relative likelihood of the token across the two regimes.

\subsection{Contrastive token-level advantage}

\method{} defines the token-level advantage as the log-likelihood contrast between light- and heavy-thinking contexts:
\begin{equation}
    A_t=\ell_t^{\lt}-\ell_t^{\htmode} = \log \pi_\phi(y_t\mid x^{\lt},y_{<t}) - \log \pi_\phi(y_t\mid x^{\htmode},y_{<t}).
\end{equation}
A positive value ($A_t > 0$) indicates that, conditioned on the shared student's prefix, the teacher considers $y_t$ more aligned with the light-thinking context, whereas a negative value indicates the converse. To stabilize policy updates, we apply two-sided clipping to bound the advantage scale. Formally, the effective advantage is:
\begin{equation}
\tilde A_t=
\operatorname{stopgrad}\!\left(
\operatorname{clip}(A_t,-A_{\max},A_{\max})
\right),\qquad A_{\max}>0.
\end{equation}
Importantly, this advantage signal incorporates no explicit length penalty. Even tokens appearing late in the sequence can receive high positive advantages, provided the teacher determines that they effectively advance concise reasoning. Consequently, reasoning compression emerges as a natural outcome of policy optimization: training reinforces tokens that make progress, summarize, or directly deliver the answer, while penalizing those that prolong redundant heavy-thinking.

\subsection{Policy update}

Given the clipped contrastive advantage $\tilde A_t$, \method{} optimizes the student policy via a clipped surrogate objective \citep{schulman2017ppo}. Let
\[
r_t(\theta)=
\frac{\pi_\theta(y_t\mid x,y_{<t})}
{\pi_{\theta_{\mathrm{old}}}(y_t\mid x,y_{<t})}
\]
denote the probability ratio between the current student policy and the old student policy used to generate the sampled responses. The clipped objective is formulated as:
\begin{equation}
J_{\mathrm{\method{}}}(\theta)=
\mathbb E_{\substack{x\sim\mathcal D\\
y\sim\pi_{\theta_{\mathrm{old}}}(\cdot\mid x)}}
\left[
\frac{1}{T_y}\sum_{t=1}^{T_y}
\min\left(
r_t(\theta)\tilde A_t,\,
\operatorname{clip}\!\left(
r_t(\theta),
1-\epsilon_{\mathrm{low}},1+\epsilon_{\mathrm{high}}
\right)\tilde A_t
\right)
\right].
\end{equation}
This objective increases the likelihood of tokens aligned with the light-thinking context while penalizing those favored under the heavy-thinking context. In this formulation, the contrastive advantage $\tilde A_t$ functions as a scalar weight that modulates the policy gradient, directly steering the optimization of the student parameters $\theta$ according to the teacher's relative preference. The overall training procedure is summarized in Algorithm~\ref{alg:copd}.

\begin{algorithm}[t]
\caption{\method{} training procedure}
\label{alg:copd}
\begin{algorithmic}[1]
\Require Initial student $\pi_{\theta_0}$, frozen teacher $\pi_\phi$, training distribution $\mathcal D$, batch size $B$, training steps $K$, advantage bound $A_{\max}$, PPO clipping bounds $\epsilon_{\mathrm{low}}$ and $\epsilon_{\mathrm{high}}$
\For{$k=0,\ldots,K-1$}
    \State $\theta_{\mathrm{old}}\gets\theta_k$
    \State Sample $\mathcal B_k=(x_i)_{i=1}^{B}$, where $x_i\overset{\mathrm{i.i.d.}}{\sim}\mathcal D$
    \For{$i=1,\ldots,B$}
        \State Sample $y_i=(y_{i,1},\ldots,y_{i,T_i})\sim\pi_{\theta_{\mathrm{old}}}(\cdot\mid x_i)$
        \State Construct $x_i^m=[\iota^m;x_i]$ for $m\in\{\lt,\htmode\}$
        \For{$t=1,\ldots,T_i$}
            \State Compute $\ell_{i,t}^m=\log\pi_\phi(y_{i,t}\mid x_i^m,y_{i,<t})$ for $m\in\{\lt,\htmode\}$
            \State $A_{i,t}\gets\ell_{i,t}^{\lt}-\ell_{i,t}^{\htmode}$
            \State $\tilde A_{i,t}\gets\operatorname{stopgrad}\!\left(\operatorname{clip}(A_{i,t},-A_{\max},A_{\max})\right)$
        \EndFor
    \EndFor
    \State Update $\theta_{k+1}\approx\argmax_{\theta}J_{\mathrm{\method{}}}(\theta)$ using $\mathcal B_k$ and the clipped ratios $[1-\epsilon_{\mathrm{low}},1+\epsilon_{\mathrm{high}}]$
\EndFor
\end{algorithmic}
\end{algorithm}

\subsection{Self-distillation variant}

This paired contrastive mechanism also applies without relying on an external teacher. We denote this variant as \selfmethod{}, where the frozen reference policy is instantiated as a static snapshot of the student model's initial weights. Beyond its practical utility, \selfmethod{} allows us to investigate whether the compression signal inherently hinges on teacher scale. If a single base model assigns distinct likelihoods to the identical action under light- versus heavy-thinking contexts, the supervisory signal is driven by context-induced relative preferences rather than external teacher capacity alone.

\section{Experiments}
\label{sec:experiments}

\subsection{Setup}

\subsubsection{Models and Baseline}

We conduct our primary experiments on Qwen3-VL-2B-Instruct and Qwen3.5-2B-Base \citep{qwen3vl2025,qwen35models2026}. For training, we select approximately 16K instances from ViRL-39K \citep{wang2025vlrethinker}, denoted as ViRL-16K\footnote{ViRL-39K reports empirical pass rates estimated via Qwen2.5-VL-32B \citep{bai2025qwen25vl}. To focus on non-trivial yet solvable instances, we retain only samples with pass rates strictly between 0 and 1. We then split the resulting 18K samples at a 9:1 ratio, yielding ViRL-16K for training and 2K for validation.}. We benchmark our method against the respective base models, Vanilla OPD \citep{lu2025onpolicydistillation}, and two competitive OPD variants: ExOPD \citep{exopd2026} and VA-OPD \citep{vaopd2026}. Specifically, ExOPD formulates online distillation as reinforcement learning under dense KL constraints, applying reward extrapolation to push student performance beyond the teacher's ceiling. VA-OPD measures the teacher's token-level log-probability gap between inputs with versus without visual details, hierarchically reweighting rollouts and tokens during distillation to prevent visual information from being diluted by linguistic tokens.

\subsubsection{Training Details}

We implement our training pipeline within \texttt{verl}, using FSDP and vLLM for accelerated on-policy rollout generation and teacher scoring \citep{sheng2024hybridflow,zhao2023fsdp,kwon2023vllm}. For each on-policy response, the frozen teacher evaluates generated tokens under contrasting light- and heavy-thinking prompts, with the resulting advantage $\ell_t^{\lt}-\ell_t^{\htmode}$ clipped to $[-10, 10]$. Crucially, we incorporate no external task rewards, KL divergence penalties, or explicit length regularizers. Models are trained for 3 epochs using the AdamW optimizer. Complete optimization hyperparameters and sequence-length configurations are provided in Appendix~\ref{app:training_details}.

\subsubsection{Evaluation Benchmarks}

We evaluate on nine multimodal benchmarks, spanning four reasoning tasks (MathVista \citep{mathvista2024}, MathVerse \citep{mathverse2024}, WeMath \citep{wemath2024}, LogicVista \citep{logicvista2024}) and five understanding tasks (HalluBench \citep{hallusionbench2024}, MMMU \citep{mmmu2024}, MMStar \citep{mmstar2024}, RWQA \citep{rwqa2024}, MMGist \citep{yuan2026mmgist})\footnote{Benchmark details and evaluation configurations are provided in the Appendix~\ref{app:benchmarks} and \ref{app:evalprotocol}.}. Table entries are formatted as Accuracy / Average Tokens / \accperk{}, where accuracy is reported as $\text{mean}_{\pm\mathrm{std}}$ over three independent runs, and response length denotes the mean. To quantify efficiency, \accperk{} measures the accuracy per thousand output tokens across the benchmark set $\mathcal{B}$:
\begin{equation}
\operatorname{\accperk{}} = 1000 \cdot \frac{\sum_{b \in \mathcal B} \operatorname{Acc}_b}{\sum_{b \in \mathcal B} \operatorname{Tokens}_b},
\end{equation}
where $\operatorname{Acc}_b$ and $\operatorname{Tokens}_b$ denote the accuracy and average response length on benchmark $b$.

\begin{table*}[t]
\centering
\caption{\textbf{Main results of \method{}.} Each benchmark cell reports Accuracy / Average Tokens; AVG reports Accuracy / Average Tokens / \accperk{}, with the standard deviation of accuracy shown as a subscript. The best accuracy, the shortest token length and the best  \accperk{} are in bold.}
\label{tab:main}
\setlength{\tabcolsep}{3pt}
\resizebox{\textwidth}{!}{
\begin{tabular}{lcccccccccc}
\toprule
\multirow{2}{*}{\textbf{Method}} & \multicolumn{4}{c}{\textbf{Reasoning}} & \multicolumn{5}{c}{\textbf{Understanding}} & \multirow{2}{*}{\textbf{AVG}} \\
\cmidrule(lr){2-5}\cmidrule(lr){6-10}
& \textbf{MathVista} & \textbf{MathVerse} & \textbf{WeMath} & \textbf{LogicVista} & \textbf{HalluBench} & \textbf{MMMU} & \textbf{MMStar} & \textbf{RWQA} & \textbf{MMGist} \\
\midrule
Q3VL-2B & 61.1/1.22K & 47.9/2.13K & 58.9/3.54K & 38.7/4.45K & 69.9/0.52K & 45.6/3.08K & 58.7/1.19K & 67.5/0.17K & 27.7/2.79K & \ms{52.9}{0.4}/2.12K/25.0 \\
\midrule
\multicolumn{11}{c}{\textbf{Qwen3-VL-8B$\to$Qwen3-VL-2B} (Avg Performance of Qwen3-VL-8B-Instruct with LT prompt: 62.9/0.55K/114.4)} \\
\midrule
OPD & 61.1/1.30K & 49.7/2.55K & 59.8/3.74K & 40.0/4.49K & 70.4/0.58K & 46.2/3.14K & 59.4/1.25K & \textbf{67.0}/0.21K & 28.0/2.86K & \ms{53.5}{0.1}/2.23K/24.0 \\
ExOPD & \textbf{66.8}/1.34K & 52.2/2.92K & 64.4/2.99K & 41.6/3.70K & \textbf{71.2}/0.50K & 47.0/2.95K & 59.5/1.27K & 66.3/0.31K & 27.9/2.79K & \ms{55.2}{0.1}/2.08K/26.5 \\
VA-OPD & 61.5/1.33K & 48.8/2.46K & 60.5/3.59K & 40.5/4.26K & 69.8/0.56K & 46.8/3.06K & 58.1/1.24K & 66.7/0.20K & 27.5/2.82K & \ms{53.4}{0.1}/2.17K/24.6 \\
\textbf{\method{}} & 62.9/\textbf{0.42K} & \textbf{53.0}/\textbf{0.73K} & \textbf{69.5}/\textbf{1.36K} & \textbf{48.4}/\textbf{1.89K} & 68.1/\textbf{0.15K} & \textbf{52.6}/\textbf{1.67K} & \textbf{60.3}/\textbf{0.47K} & \textbf{67.0}/\textbf{0.03K} & \textbf{31.2}/\textbf{1.42K} & \msb{57.0}{0.2}/\textbf{0.91K}/\textbf{62.6} \\
\midrule
\multicolumn{11}{c}{\textbf{Qwen3-VL-32B$\to$Qwen3-VL-2B} (Avg Performance of Qwen3-VL-32B-Instruct with LT prompt: 72.0/0.69K/104.3)} \\
\midrule
OPD & 63.4/1.34K & 50.8/2.92K & 62.1/3.77K & 42.1/4.38K & \textbf{70.4}/0.63K & 47.3/3.07K & 60.5/1.25K & 67.5/0.25K & 28.9/2.92K & \ms{54.8}{0.3}/2.28K/24.0 \\
ExOPD & \textbf{66.9}/1.20K & 53.9/2.68K & 65.7/2.82K & 44.4/3.65K & 67.9/0.35K & 48.9/2.66K & \textbf{63.6}/0.99K & 68.1/0.28K & 30.3/2.56K & \ms{56.6}{0.2}/1.91K/29.6 \\
VA-OPD & 63.7/1.34K & 50.2/2.87K & 63.0/3.57K & 42.9/4.19K & 69.9/0.58K & 47.7/2.97K & 60.4/1.24K & \textbf{68.5}/0.27K & 28.8/2.85K & \ms{55.0}{0.2}/2.21K/24.9 \\
\textbf{\method{}} & 62.9/\textbf{0.39K} & \textbf{54.0}/\textbf{0.71K} & \textbf{71.6}/\textbf{1.16K} & \textbf{49.4}/\textbf{1.51K} & 66.6/\textbf{0.15K} & \textbf{53.2}/\textbf{1.44K} & 61.0/\textbf{0.41K} & 64.9/\textbf{0.03K} & \textbf{31.1}/\textbf{1.23K} & \msb{57.2}{0.1}/\textbf{0.78K}/\textbf{73.3} \\
\midrule
\midrule
Q35-2B & 68.8/2.25K & 55.3/3.92K & 60.9/4.18K & 43.2/4.23K & 69.9/0.99K & 48.5/3.54K & 62.4/1.90K & 69.1/1.05K & 30.3/3.47K & \ms{56.5}{0.1}/2.84K/19.9 \\
\midrule
\multicolumn{11}{c}{\textbf{Qwen3.5-4B$\to$Qwen3.5-2B} (Avg Performance of Qwen3.5-4B with LT prompt: 72.5/2.67K/27.2)} \\
\midrule
OPD & 70.2/2.95K & 57.5/4.55K & 62.3/5.09K & 40.1/5.63K & 70.4/1.73K & 49.7/4.31K & 62.2/2.48K & 67.7/1.81K & 29.4/4.47K & \ms{56.6}{0.4}/3.67K/15.4 \\
ExOPD & 69.8/3.08K & 58.6/4.56K & 62.6/5.14K & 41.3/5.58K & 71.7/1.73K & 49.9/4.24K & 63.6/2.53K & \textbf{68.7}/1.74K & 29.0/4.48K & \ms{57.3}{0.5}/3.68K/15.6 \\
VA-OPD & 68.0/3.10K & 56.9/4.60K & 61.2/5.19K & 41.0/5.45K & 70.9/1.76K & 49.2/4.28K & 62.7/2.56K & 68.3/1.74K & 28.3/4.46K & \ms{56.3}{0.3}/3.68K/15.3 \\
\textbf{\method{}} & \textbf{71.2}/\textbf{1.49K} & \textbf{60.9}/\textbf{2.37K} & \textbf{68.9}/\textbf{2.64K} & \textbf{49.3}/\textbf{3.06K} & \textbf{71.9}/\textbf{0.63K} & \textbf{54.4}/\textbf{2.65K} & \textbf{64.9}/\textbf{1.29K} & 68.3/\textbf{0.68K} & \textbf{31.7}/\textbf{2.57K} & \msb{60.2}{0.1}/\textbf{1.93K}/\textbf{31.2} \\
\midrule
\multicolumn{11}{c}{\textbf{Qwen3.5-35A3B$\to$Qwen3.5-2B} (Avg Performance of Qwen3.5-35A3B with LT prompt: 78.5/2.58K/30.4)} \\
\midrule
OPD & 69.2/3.15K & 58.6/4.62K & 62.7/5.22K & 39.5/5.69K & \textbf{71.5}/1.86K & 50.1/4.44K & 62.5/2.68K & 65.7/2.01K & 28.7/4.67K & \ms{56.5}{0.1}/3.82K/14.8 \\
ExOPD & 69.7/3.14K & 58.8/4.60K & 63.8/5.21K & 39.9/5.56K & 70.3/1.82K & 50.2/4.35K & 63.8/2.63K & 67.1/1.94K & 28.0/4.63K & \ms{56.8}{0.1}/3.76K/15.1 \\
VA-OPD & 69.2/3.21K & 57.7/4.60K & 62.0/5.24K & 42.8/5.54K & 71.2/1.81K & 50.1/4.39K & 62.6/2.62K & 67.1/1.85K & 27.7/4.60K & \ms{56.7}{0.2}/3.76K/15.1 \\
\textbf{\method{}} & \textbf{70.8}/\textbf{1.59K} & \textbf{60.1}/\textbf{2.54K} & \textbf{68.0}/\textbf{2.86K} & \textbf{48.4}/\textbf{3.24K} & 71.1/\textbf{0.62K} & \textbf{52.9}/\textbf{2.85K} & \textbf{64.6}/\textbf{1.38K} & \textbf{69.5}/\textbf{0.70K} & \textbf{31.3}/\textbf{2.73K} & \msb{59.6}{0.4}/\textbf{2.06K}/\textbf{28.9} \\
\bottomrule
\end{tabular}
}
\end{table*}

\subsection{Main Results}

\noindent\textbf{Accuracy--length Pareto frontier.} Table~\ref{tab:main} demonstrates that \method{} establishes a superior accuracy--length Pareto frontier across all four teacher--student configurations, achieving the highest average accuracy alongside the shortest response lengths. For Qwen3-VL-8B$\to$2B, relative to the strong baseline ExOPD, \method{} improves accuracy by 1.8 pp while cutting response length by 56.3\%, thereby increasing \accperk{} by 136.2\%. This advantage consistently holds across the remaining model and teacher pairings, where \method{} outperforms ExOPD with a 91.4\%--147.6\% boost in \accperk{}. These results confirm that \method{} steers the student policy toward concise reasoning behaviors while achieving better performance.

\noindent\textbf{Broad compression with task-level heterogeneity.} Under the Qwen3-VL-8B$\to$2B configuration, relative to the un-distilled Qwen3-VL-2B base model, \method{} reduces generation length across all nine benchmarks while improving accuracy on seven. The most pronounced gains appear on complex reasoning benchmarks, namely WeMath (+10.6 pp accuracy with 61.6\% fewer tokens) and LogicVista (+9.7 pp accuracy with 57.5\% fewer tokens), demonstrating that aggressive compression does not compromise essential reasoning steps on tasks with inherently verbose initial traces. In contrast, on HalluBench and RWQA where initial responses are already brief, accuracy decreases slightly by 1.8 pp and 0.5 pp alongside substantial token reductions of 71.2\% and 82.4\%, respectively. This pattern indicates that while the contrastive advantage effectively eliminates redundant rationales, it delineates an operational boundary: aggressive compression provides a narrower safety margin when tasks inherently elicit concise responses.

\noindent\textbf{Gains do not follow monotonic teacher scaling.} Scaling the Qwen3-VL teacher from 8B to 32B improves average accuracy by 0.2 pp, reduces response length by 14.3\%, and boosts \accperk{} by 17.1\%. Conversely, scaling the Qwen3.5 teacher from 4B to 35A3B decreases accuracy by 0.6 pp, lengthens responses by 6.7\%, and lowers \accperk{} by 7.4\%. Consequently, increasing teacher scale shifts the trade-off frontier without reliably optimizing it. This non-monotonicity is consistent with recent findings in on-policy distillation: simply deploying a larger or higher-scoring teacher does not necessarily yield more effective token-level supervisory signals \citep{rethinkingopd2026}. Nevertheless, \method{} achieves the highest \accperk{} and shortest average responses across all four teacher--student configurations, confirming that our consistent gains stem primarily from the contrastive preferences extracted from the teacher, rather than teacher capacity alone.

\subsection{Ablation Studies}

In this section, we systematically conduct ablation studies across multiple dimensions. Unless noted otherwise, all experiments adhere to the same training and evaluation configurations as the main experiments (see Appendices~\ref{app:mmesci}--\ref{app:words} for more experimental results).

\subsubsection{Generalization on Scales}

\begin{table}[t]
\centering
\caption{\textbf{Generalization across various model scales.}}
\label{tab:scaling}
\resizebox{0.95\textwidth}{!}{
\begin{tabular}{lccc}
\toprule
\textbf{Method} & \textbf{Reasoning} & \textbf{Understanding} & \textbf{AVG} \\
\midrule
Qwen3.5-4B & \ms{72.2}{0.1}/2.87K/25.2 & \ms{64.9}{0.1}/1.85K/35.1 & \ms{68.1}{0.1}/2.30K/29.6 \\
\textbf{Qwen3.5-4B $\xleftarrow{\text{\method{}}}$ Qwen3.5-35A3B} & \msb{73.9}{0.2}/\textbf{1.52K}/\textbf{48.6} & \msb{65.2}{0.5}/\textbf{1.22K}/\textbf{53.4} & \msb{69.1}{0.2}/\textbf{1.35K}/\textbf{51.2} \\
\midrule
Qwen3.5-9B & \ms{75.8}{0.4}/2.73K/27.8 & \ms{67.6}{0.4}/1.85K/36.5 & \ms{71.2}{0.2}/2.24K/31.8 \\
\textbf{Qwen3.5-9B $\xleftarrow{\text{\method{}}}$ Qwen3.5-35A3B} & \ms{76.8}{0.3}/\textbf{1.52K}/\textbf{50.5} & \ms{68.1}{0.1}/\textbf{1.29K}/\textbf{52.8} & \ms{72.0}{0.1}/\textbf{1.40K}/\textbf{51.4} \\
\textbf{Qwen3.5-9B $\xleftarrow{\text{\method{}}}$ Qwen3.5-122A10B} & \msb{78.6}{0.4}/1.99K/39.5 & \msb{68.8}{0.5}/1.56K/44.1 & \msb{73.2}{0.4}/1.75K/41.8 \\
\bottomrule
\end{tabular}
}
\end{table}

As shown in Table~\ref{tab:scaling}, the contrastive signal generalizes effectively across different student scales while maintaining its compression efficacy. Holding the 35A3B teacher fixed, \method{} shortens average response length by 41.3\% and 37.5\%, improves average accuracy by 1.0 pp and 0.8 pp, and boosts \accperk{} by 73.0\% and 61.6\% on the 4B and 9B students, respectively. Conversely, fixing the 9B student and scaling the teacher from 35A3B to 122A10B yields an additional 1.2 pp in accuracy, but at the cost of a 25.0\% increase in response length, which reduces \accperk{} by 18.7\%. These findings indicate that teacher scaling primarily shifts the accuracy--efficiency trade-off, whereas the efficiency benefits of \method{} persist consistently across all evaluated student scales.

\subsubsection{Generalization Across Training Data}

\begin{table}[t]
\centering
\caption{\textbf{Generalization across training data under the Qwen3-VL-2B $\xleftarrow{\text{\method{}}}$ Qwen3-VL-8B setting.} Geo3K (24 epochs) is iteration-matched to ViRL-16K (3 epochs).}
\label{tab:trainingdata}
\resizebox{0.85\textwidth}{!}{
\begin{tabular}{llccc}
\toprule
\textbf{Dataset} & \textbf{Epoch} & \textbf{Reasoning} & \textbf{Understanding} & \textbf{AVG} \\
\midrule
- & -  &\ms{51.6}{0.6}/2.84K/18.2 & \ms{53.9}{0.2}/1.55K/34.8 & \ms{52.9}{0.4}/2.12K/25.0 \\
Geo3K & 3 & \ms{54.3}{0.1}/2.11K/25.7 & \ms{54.4}{0.1}/1.25K/43.5 & \ms{54.3}{0.1}/1.63K/33.3 \\
Geo3K & 24  &\ms{58.1}{0.3}/1.24K/46.9 & \ms{55.2}{0.1}/0.83K/66.5 & \ms{56.5}{0.1}/1.01K/55.9 \\
ViRL-16K & 3 & \msb{58.5}{0.3}/\textbf{1.10K}/\textbf{53.2} & \msb{55.8}{0.2}/\textbf{0.75K}/\textbf{74.4} & \msb{57.0}{0.2}/\textbf{0.91K}/\textbf{62.6} \\
\bottomrule
\end{tabular}
}
\end{table}

Table~\ref{tab:trainingdata} compares \method{} trained on Geo3K \citep{lu2021intergpsinterpretablegeometryproblem} against ViRL-16K. Relative to the baseline, Geo3K (3 epochs) increases average accuracy from 52.9\% to 54.3\% while reducing average response length by 23.1\%. Extending Geo3K training to 24 epochs yields 56.5\% accuracy at 1.01K tokens, elevating \accperk{} to 55.9. In comparison, ViRL-16K (3 epochs) achieves 57.0\% accuracy at 0.91K tokens, attaining the highest \accperk{} of 62.6. These results demonstrate that \method{} does not depend on a specialized data distribution, consistently improving the accuracy--length trade-off across both domain-specific and general multimodal corpora.

\subsubsection{Generalization on Various Prompts}

\begin{table}[t]
\centering
\caption{\textbf{Robustness on various LT/HT prompts using Qwen3-VL-8B as the teacher.}}
\label{tab:variousprompts}
\resizebox{0.85\textwidth}{!}{
\begin{tabular}{lccc}
\toprule
\textbf{Method} & \textbf{Reasoning} & \textbf{Understanding} & \textbf{AVG} \\
\midrule
Qwen3-VL-2B & \ms{51.6}{0.6}/2.84K/18.2 & \ms{53.9}{0.2}/1.55K/34.8 & \ms{52.9}{0.4}/2.12K/25.0 \\
Simple & \ms{58.2}{0.5}/1.12K/52.0 & \ms{55.3}{0.3}/0.76K/72.8 & \ms{56.6}{0.3}/0.92K/61.5 \\
Complex & \ms{57.6}{0.5}/\textbf{0.97K}/\textbf{59.4} & \ms{55.0}{0.2}/\textbf{0.70K}/\textbf{78.6} & \ms{56.2}{0.3}/\textbf{0.82K}/\textbf{68.5} \\
\textbf{\method{} (Ours)} & \msb{58.5}{0.3}/1.10K/53.2 & \msb{55.8}{0.2}/0.75K/74.4 & \msb{57.0}{0.2}/0.91K/62.6 \\
\bottomrule
\end{tabular}
}
\end{table}

As shown in Table~\ref{tab:variousprompts}, \method{} is robust to prompt phrasing: the Simple, Complex, and default templates achieve comparable accuracy (56.2\%--57.0\%), response length (0.82--0.92K tokens), and \accperk{} (61.5--68.5), all consistently outperforming the base model. This consistency confirms that the gains stem from the contrastive formulation itself rather than delicate prompt engineering.

\subsubsection{Contrastive On-Policy Self-Distillation}

\begin{table}[t]
\centering
\caption{\textbf{Contrastive on-policy self-distillation.} \selfmethod{} uses a frozen copy of the initial student model as the paired LT--HT scorer, eliminating the need for an external teacher.}
\label{tab:copsd}
\resizebox{0.85\textwidth}{!}{
\begin{tabular}{lccc}
\toprule
\textbf{Method} & \textbf{Reasoning} & \textbf{Understanding} & \textbf{AVG} \\
\midrule
Qwen3-VL-2B & \ms{51.6}{0.6}/2.84K/18.2 & \ms{53.9}{0.2}/1.55K/34.8 & \ms{52.9}{0.4}/2.12K/25.0 \\
\textbf{+\selfmethod{}} & \msb{58.3}{0.5}/\textbf{0.92K}/\textbf{63.4} & \msb{55.4}{0.2}/\textbf{0.64K}/\textbf{86.6} & \msb{56.7}{0.3}/\textbf{0.76K}/\textbf{74.6} \\
\midrule
Qwen3-VL-4B & \msb{69.3}{0.4}/1.56K/44.4 & \ms{63.0}{0.2}/1.03K/61.2 & \ms{65.8}{0.2}/1.26K/52.2 \\
\textbf{+\selfmethod{}} & \ms{69.0}{0.3}/\textbf{0.65K}/\textbf{106.2} & \msb{63.4}{0.1}/\textbf{0.53K}/\textbf{119.6} & \msb{65.9}{0.1}/\textbf{0.58K}/\textbf{113.6} \\
\midrule
Qwen3-VL-8B & \msb{71.0}{0.6}/1.41K/50.4 & \ms{64.4}{0.2}/0.92K/70.0 & \msb{67.3}{0.4}/1.14K/59.0 \\
\textbf{+\selfmethod{}} & \ms{70.8}{0.0}/\textbf{0.57K}/\textbf{124.2} & \msb{64.5}{0.0}/\textbf{0.48K}/\textbf{134.4} & \msb{67.3}{0.0}/\textbf{0.52K}/\textbf{129.4} \\
\midrule
Qwen3.5-122A10B & \ms{84.4}{0.1}/3.81K/22.2 & \ms{75.1}{0.3}/3.10K/24.2 & \ms{79.2}{0.3}/3.42K/23.2 \\
\textbf{+\selfmethod{}} & \msb{86.1}{0.1}/\textbf{3.14K}/\textbf{27.4} & \msb{76.3}{0.1}/\textbf{2.55K}/\textbf{29.9} & \msb{80.7}{0.0}/\textbf{2.81K}/\textbf{28.7} \\
\bottomrule
\end{tabular}
}
\end{table}

While larger external teachers provide benefits, they are not strictly necessary for acquiring concise reasoning behaviors. Results across four model scales in Table~\ref{tab:copsd} demonstrate that a frozen snapshot of the student itself suffices to construct effective contrastive signals. For the 2B model, \selfmethod{} improves average accuracy by 3.8 pp while reducing response length by 64.2\%, resulting in a 198.4\% boost in \accperk{}. The same trend consistently emerges across the 4B, 8B, and 122A10B scales: average accuracy is maintained or improved, response length shrinks by 17.8\%--54.4\%, and \accperk{} increases by 23.7\%--119.3\%. These findings indicate that effective compression does not strictly hinge on teacher scale. Instead, the model's own relative preference between light- and heavy-thinking contexts provides sufficient supervision to steer generation toward concise outputs.

The gains of \selfmethod{} reflect a more effective use of the model's existing reasoning capabilities through policy adaptation. Prior work demonstrates that language models possess a latent capacity for concise reasoning, which can be elicited through self-training on their own generated trajectories while preserving accuracy \citep{munkhbat2025concise}. Moreover, reasoning accuracy does not increase monotonically with chain-of-thought length: extending reasoning beyond an appropriate length can reduce performance \citep{wu2026whenmore}. Analyses of reasoning traces further identify redundant repetition, excessive self-verification, and over-exploration of alternatives as common sources of unnecessary computation \citep{yi2025shorterbetter}. In \selfmethod{}, the LT and HT scoring contexts expose different conditional preferences within the same frozen model. Optimizing their likelihood contrast reinforces student-generated continuations favored by the light-thinking context, shifting the generation policy toward more concise responses.

\subsubsection{Can a single light-thinking teacher foster light-thinking in students?}

\begin{table}[t]
\centering
\caption{\textbf{Ablation of the contrastive teacher signal with Qwen3-VL-8B as teacher.} Using only the light-thinking log-probability does not yield output-efficient reasoning, whereas the LT--HT contrast improves accuracy while reducing response length.}
\label{tab:onlylt}
\resizebox{0.85\textwidth}{!}{
\begin{tabular}{lccc}
\toprule
\textbf{Method} & \textbf{Reasoning} & \textbf{Understanding} & \textbf{AVG} \\
\midrule
Qwen3-VL-2B & \ms{51.6}{0.6}/2.84K/18.2 & \ms{53.9}{0.2}/1.55K/34.8 & \ms{52.9}{0.4}/2.12K/25.0 \\
$A_t=\ell_t^{\lt}$ & \ms{52.4}{0.6}/3.05K/17.2 & \ms{53.9}{0.4}/1.62K/33.3 & \ms{53.2}{0.1}/2.25K/23.6 \\
$A_t=\ell_t^{\lt}-\ell_t^{\htmode}$ & \msb{58.5}{0.3}/\textbf{1.10K}/\textbf{53.2} & \msb{55.8}{0.2}/\textbf{0.75K}/\textbf{74.4} & \msb{57.0}{0.2}/\textbf{0.91K}/\textbf{62.6} \\
\bottomrule
\end{tabular}
}
\end{table}

The critical factor driving compression is the contrast between LT and HT prompts, rather than the LT score alone. Results in Table~\ref{tab:onlylt} illustrate this distinction: optimizing solely for the LT score lengthens outputs, whereas the contrastive objective improves both accuracy and efficiency. Specifically, adopting a unilateral objective $A_t=\ell_t^{\lt}$ yields only a marginal gain in average accuracy (52.9\% $\to$ 53.2\%) while expanding average response length from 2.12K to 2.25K tokens. These results indicate that conditional likelihood under LT alone fails to establish a definitive criterion for determining which reasoning steps to streamline. In contrast, the formulation $A_t=\ell_t^{\lt}-\ell_t^{\htmode}$ evaluates each token under both contexts, converting teacher evaluations into a directional signal actively favoring concise reasoning. This contrastive objective elevates average accuracy to 57.0\% while slashing average response length by 57.1\%. These findings confirm that the HT score provides an indispensable baseline, enabling the contrastive signal to selectively penalize redundant continuations rather than merely imitate the teacher distribution under a LT prompt.

\subsubsection{Comparison with Other Training Paradigms}

\begin{table}[t]
\centering
\caption{\textbf{Comparison with other training paradigms.} We use Qwen3-VL-8B as the teacher and Qwen3-VL-2B as the student. \textbf{Note that when evaluated under LT and HT prompts, Qwen3-VL-8B achieves an average performance of 62.9/0.55K/114.4 and 66.3/1.94K/34.2, respectively.}}
\label{tab:trainingparadigm}
\resizebox{0.9\textwidth}{!}{
\begin{tabular}{lccc}
\toprule
\textbf{Method} & \textbf{Reasoning} & \textbf{Understanding} & \textbf{AVG} \\
\midrule
Qwen3-VL-2B & \ms{51.6}{0.6}/2.84K/18.2 & \ms{53.9}{0.2}/1.55K/34.8 & \ms{52.9}{0.4}/2.12K/25.0 \\
\quad + PE with $\lt$ & \ms{53.1}{0.1}/2.00K/26.6 & \ms{53.3}{0.4}/1.04K/51.3 & \ms{53.2}{0.3}/1.47K/36.2 \\
\quad + SFT on $\lt$ data & \ms{52.6}{0.4}/2.44K/21.6 & \ms{52.9}{0.4}/1.37K/38.6 & \ms{52.8}{0.2}/1.85K/28.5 \\
\quad + DPO ($\lt$ data vs. $\htmode$ data) & \ms{53.8}{0.7}/1.99K/27.0 & \ms{53.0}{0.1}/1.15K/46.1 & \ms{53.4}{0.3}/1.53K/34.9 \\
\textbf{\quad + \method{} (Ours)} & \msb{58.5}{0.3}/\textbf{1.10K}/\textbf{53.2} & \msb{55.8}{0.2}/\textbf{0.75K}/\textbf{74.4} & \msb{57.0}{0.2}/\textbf{0.91K}/\textbf{62.6} \\
\bottomrule
\end{tabular}
}
\end{table}

Table~\ref{tab:trainingparadigm} demonstrates that alternative training paradigms fail to fully reproduce the benefits of our LT--HT contrastive formulation. Specifically, we use Qwen3-VL-8B as the teacher to generate offline responses under the light- and heavy-thinking instructions for SFT and DPO \citep{rafailov2023direct} training. Among the evaluated alternatives, DPO achieves the highest average accuracy at 53.4\%, whereas prompt engineering (PE)---which directly applies the light-thinking instruction to the student model during inference---yields the shortest average response (1.47K tokens) and highest \accperk{} (36.2). In comparison, \method{} advances average accuracy to 57.0\% (+3.6 pp over DPO), while simultaneously cutting response length by 38.1\% and boosting \accperk{} by 72.9\% relative to PE. These findings demonstrate that an on-policy, token-level contrastive advantage provides far more granular supervision and credit assignment than sequence-level imitation or offline preference optimization, suppressing redundant reasoning without compromising problem-solving accuracy.

\subsection{Training Efficiency}

\begin{table}[!t]
\centering
\begin{minipage}{0.48\linewidth}
\centering
\caption{\textbf{Training efficiency under the Qwen3-VL-8B$\to$2B.} GPU hours denote the aggregate training time across all GPUs.}
\label{tab:trainingtime}
\small
\setlength{\tabcolsep}{5pt}
\begin{tabular}{@{}lccc@{}}
\toprule
\textbf{Method} & \textbf{GPU Hours} & \textbf{AVG} \\
\midrule
Base & -- & \ms{52.9}{0.4}/2.12K/25.0 \\
OPD & 132 & \ms{53.5}{0.1}/2.23K/24.0 \\
ExOPD & 150 & \ms{55.2}{0.1}/2.08K/26.5 \\
\textbf{\method{}} & \textbf{60} & \msb{57.0}{0.2}/\textbf{0.91K}/\textbf{62.6} \\
\bottomrule
\end{tabular}
\end{minipage}%
\hfill
\begin{minipage}{0.48\linewidth}
\centering
\includegraphics[width=\linewidth]{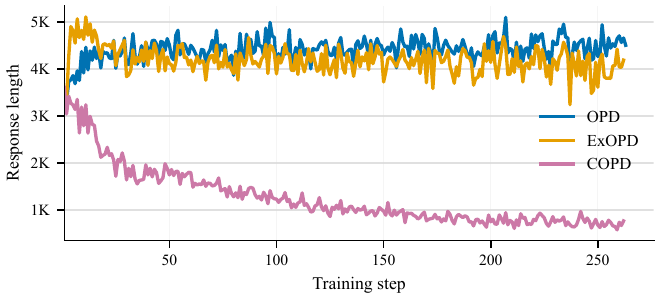}
\captionof{figure}{\textbf{Response length over training.} \method{} progressively shortens rollout responses, whereas others remain at high response lengths.}
\label{fig:resplen}
\end{minipage}
\end{table}

Table~\ref{tab:trainingtime} and Figure~\ref{fig:resplen} demonstrate that \method{} excels in both training and inference efficiency. It requires merely 60 GPU hours, reducing compute costs by 54.5\% over Vanilla OPD and 60.0\% over ExOPD. Concurrently, \method{} delivers the highest average accuracy with the shortest outputs. Crucially, its training dynamics (Figure~\ref{fig:resplen}) show a progressive reduction in response length below 1K tokens, while competing methods remain largely static.

\subsection{Case Study}
\label{sec:case_study}

\begin{figure}[!t]
    \centering
    \includegraphics[width=0.99\linewidth]{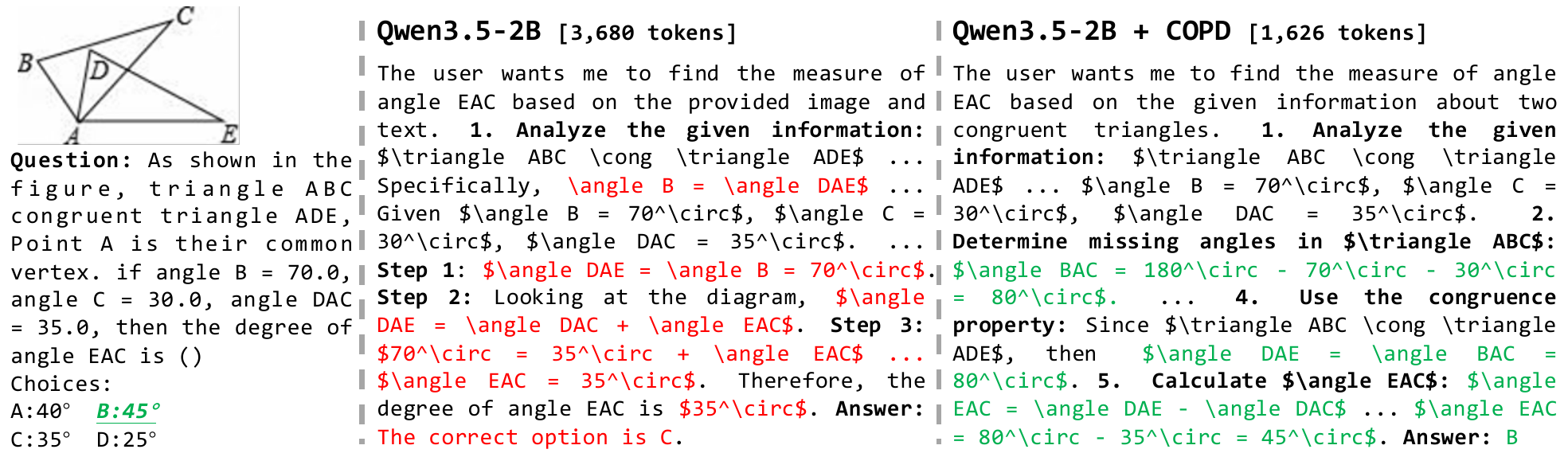}
    \caption{\textbf{Case study.} \method{} preserves decisive steps while removing redundancy.}
    \label{fig:maincase}
\end{figure}

Figure~\ref{fig:maincase} illustrates a representative case demonstrating this behavioral shift. The base model (Qwen3.5-2B) produces an overly verbose trajectory of 3,680 tokens, yet falters during repetitive self-checking and derives an incorrect answer. In contrast, \method{} reaches the correct solution directly with only 1,626 tokens (44.2\% of the base model's length). This confirms that verbose deliberation does not prevent reasoning errors; instead, \method{} effectively prunes unproductive verification loops while preserving decisive reasoning steps. More case studies can be found in Appendix~\ref{app:case_studies}.

\section{Related Work}

\noindent\textbf{On-policy distillation and multimodal reasoning.} While large vision-language models have made substantial progress across multimodal tasks through step-by-step reasoning, their rationales frequently accumulate redundant visual restatements and excessive deliberation \citep{jiang2025mme}. To enhance student capabilities, on-policy distillation (OPD) mitigates the exposure bias of offline distillation by sampling rollouts from the student policy and having a teacher score the visited prefixes \citep{lu2025onpolicydistillation}. Recent variants refine reward scaling (ExOPD \citep{exopd2026}) or visual grounding (VA-OPD \citep{vaopd2026}) to improve the student performance. In contrast, \method{} shifts the axis of supervision: the frozen teacher scores each token at the identical student state under contrasting light- and heavy-thinking prompts, providing a state-matched preference between concise and exhaustive reasoning.

\noindent\textbf{Reasoning compression and preference learning.} Prior efforts to compress reasoning chains primarily rely on supervised fine-tuning over curated concise rationales \citep{munkhbat2025concise} or reinforcement learning with explicit token budgets and heuristic length penalties \citep{han2024tokenbudget,jin2025recut,liu2025dler}. However, these approaches largely depend on scarce curated offline datasets, fragile length hyperparameters, or coarse outcome-level feedback. In contrast, \method{} requires no external task rewards or explicit length regularizers; instead, it directly derives intrinsic token-level compression preferences from paired teacher likelihoods evaluated at student-visited states. A broader discussion of related works is provided in Appendix~\ref{app:expanded_related_work}.

\section{Conclusion}

We propose \method{}, a contrastive on-policy distillation framework for adaptive reasoning compression. By evaluating each student-generated token under paired light- and heavy-thinking prompts, a frozen teacher model converts absolute token likelihoods into relative preference signals that encourage concise reasoning. This design preserves the benefits of online supervision in OPD without relying on predefined length budgets or auxiliary penalties. Extensive experiments, including evaluations on nine multimodal benchmarks across four teacher--student configurations and self-distillation variants, show that \method{} consistently improves the accuracy--efficiency Pareto frontier. These results suggest that concise reasoning can emerge from relative preferences without sacrificing problem-solving accuracy. More broadly, future work could extend this pairwise comparison mechanism to other behavioral alignment tasks, such as enforcing safety constraints, controlling writing style, and improving factuality.

\bibliography{references}
\bibliographystyle{iclr2027_conference}

\clearpage
\appendix
\section{Extended Experimental Setup and Analysis}
\label{app:additional_experiments}

\subsection{Training Infrastructure and Hyperparameters}
\label{app:training_details}

We implement our on-policy distillation pipeline within \texttt{verl} \citep{sheng2024hybridflow}, combining Fully Sharded Data Parallel (FSDP \citep{zhao2023fsdp}) for distributed policy updates with vLLM \citep{kwon2023vllm} for high-throughput rollout generation and teacher scoring. At each iteration, the student policy samples response trajectories for a batch of inputs from ViRL-16K. Conditioned on the shared multimodal input and generated response prefix, the frozen teacher scores each sampled trajectory twice under contrasting light- and heavy-thinking prompts.

Following Section~\ref{sec:method}, the token-level contrastive advantage and its clipped form are defined as:
\begin{equation}
 A_t=\ell_t^{\lt}-\ell_t^{\htmode},\qquad
 \tilde A_t=\operatorname{stopgrad}\!\left(\operatorname{clip}(A_t,-10,10)\right).
\end{equation}
This clipped advantage functions as a scalar weight in the policy-gradient objective defined in the main paper, with gradients updating the student parameters exclusively. Bounding the advantage magnitude prevents destabilizing updates and ensures numerical stability.

Regarding optimization hyperparameters, we employ the AdamW optimizer \citep{loshchilov2017adamw} with a learning rate of $1\times10^{-6}$, weight decay of $1\times10^{-2}$, and a maximum gradient clipping norm of 1.0. The global batch size is 192 with a PPO mini-batch size of 48. Training runs for 3 epochs, with maximum sequence lengths for input prompts and model responses capped at 4,096 and 8,192 tokens, respectively. The training objective incorporates no auxiliary task, format, accuracy, or explicit length rewards.

\subsection{Evaluation Benchmarks}
\label{app:benchmarks}

We evaluate on nine multimodal benchmarks, comprising a total of $18{,}656$ test instances. For structured presentation and analysis, we partition these benchmarks into four reasoning suites and five understanding suites:

\textbf{Multimodal Reasoning.}
\textsc{MathVista} \citep{mathvista2024} (\emph{testmini}, $1{,}000$ questions) benchmarks mathematical reasoning across diverse visual contexts, including natural photographs, synthetic charts, geometric diagrams, and academic figures, using both free-form generation and multiple-choice formats.
\textsc{MathVerse} \citep{mathverse2024} (\emph{testmini}, $3{,}940$ evaluation instances) evaluates plane geometry, solid geometry, and functions through five visual variants of $788$ core problems that systematically redistribute information between text and diagrams; we exclude the text-only variant to maintain strict multimodal evaluation.
\textsc{WeMath} \citep{wemath2024} (\emph{testmini}, $1{,}740$ questions) probes hierarchical visual mathematical reasoning by pairing complex multi-step problems with their prerequisite sub-problems, featuring an explicit ``No correct answer'' option to penalize superficial guesswork.
\textsc{LogicVista} \citep{logicvista2024} ($448$ multiple-choice questions) assesses deductive, inductive, numerical, spatial, and mechanical reasoning, sourcing visual questions directly from licensed intelligence and aptitude tests.

\textbf{Multimodal Understanding.}
\textsc{HallusionBench} \citep{hallusionbench2024} (\textsc{HalluBench}; $951$ image-grounded questions) diagnoses language hallucinations and visual illusions via paired yes/no queries targeting original versus human-edited images. We evaluate on the $951$ image-bearing instances from the full $1{,}129$-question benchmark, omitting the $178$ text-only controls.
\textsc{MMMU} \citep{mmmu2024} (\emph{dev}$+$\emph{val}, $1{,}050$ questions) tests college-level domain expertise across six disciplines spanning STEM, business, and the humanities.
\textsc{MMStar} \citep{mmstar2024} ($1{,}500$ questions across six core capabilities) enforces strict visual dependency through rigorous expert human verification, filtering out instances solvable via unimodal language priors.
\textsc{RealWorldQA} \citep{rwqa2024} (\textsc{RWQA}; $765$ questions) evaluates physical spatial understanding in real-world scenes.
\textsc{MMGist} \citep{yuan2026mmgist} ($7{,}262$ questions spanning seven capability dimensions) is distilled from eighteen public benchmarks via text-ablation filtering, cross-model saturation screening, and human-verified anomaly removal.

\textbf{Prompt and Evaluation Protocol.}
All primary evaluations follow a strict zero-shot protocol, where model inputs are restricted to the visual inputs, question text, and candidate options when available. For \textsc{HallusionBench}, we append a standard yes/no response formatting directive. Input formatting and response constraints remain identical across all evaluated models, without custom system prompts or in-context demonstrations.

\subsection{Evaluation Implementation Details}
\label{app:evalprotocol}

\textbf{Serving and Decoding.} We serve standalone model checkpoints via vLLM (v$0.18.0$) in bfloat16 precision. Unless specified otherwise, decoding hyperparameters are set to \texttt{max\_tokens}${}=8192$, \texttt{temperature}${}=0.7$, \texttt{top\_p}${}=0.8$, \texttt{top\_k}${}=20$, \texttt{presence\_penalty}${}=1.5$, \texttt{repetition\_penalty}${}=1.0$, and \texttt{seed}${}=3407$.

\textbf{Repeated Evaluations.} Primary results summarize three evaluation passes, with accuracy reported as $\text{mean}_{\pm\mathrm{std}}$ and response length reported as the mean.

\textbf{Automated Grading.} We employ Qwen3.6-Flash at temperature $0$ as the automated judge, prompted with the question, ground-truth answer, and candidate response to output either \texttt{correct} or \texttt{incorrect}. When a response contains a \texttt{\textbackslash boxed\{\}} expression, only the boxed content is extracted for grading. Accuracy is computed as the proportion of responses graded correct.

\textbf{Evaluation Metrics.} Sequence lengths are computed using each model's native tokenizer, measuring the complete assistant response inclusive of reasoning segments. Reasoning, Understanding, and overall AVG denote the averages across four, five, and all nine benchmarks, respectively. To measure performance density, \accperk{} calculates the accuracy per thousand output tokens across the benchmark set $\mathcal{B}$:
\begin{equation}
\operatorname{\accperk{}} = 1000 \cdot \frac{\sum_{b \in \mathcal B} \operatorname{Acc}_b}{\sum_{b \in \mathcal B} \operatorname{Tokens}_b},
\end{equation}
where $\operatorname{Acc}_b$ and $\operatorname{Tokens}_b$ denote the accuracy and average response length on benchmark $b$, respectively.

\textbf{Evaluated Variations.} To examine behavior across diverse inference settings, Figure~\ref{fig:pareto} varies the generation budget over $\{2048, 4096, 8192, 16384\}$ tokens. In addition, the text-only evaluation in Table~\ref{tab:textonlyavg8} expands the generation budget to 32K tokens, Table~\ref{tab:nothink} examines inference with thinking disabled (\texttt{enable\_thinking}=\texttt{false}), and Table~\ref{tab:temperatureis0} evaluates greedy decoding (temperature $0$).

\subsection{Generalization to High-School STEM Reasoning}
\label{app:mmesci}

\begin{table}[t]
\centering
\caption{\textbf{Scientific reasoning on MMESci-VisionOnly.} Performance across high-school Physics, Math, Chemistry, Biology, and overall evaluation (All). Each cell reports accuracy ($\text{mean}_{\pm\mathrm{std}}$), average response length (in thousands of tokens), and the efficiency index \accperk{}.}
\label{tab:mmesci}
\resizebox{\linewidth}{!}{
\begin{tabular}{lccccc}
\toprule
\textbf{Method} & \textbf{Physics} & \textbf{Math} & \textbf{Chemistry} & \textbf{Biology} & \textbf{All} \\
\midrule
Qwen3-VL-2B & \ms{6.9}{0.2}/5.00K/1.4 & \ms{13.2}{1.7}/6.27K/2.1 & \ms{14.1}{1.6}/3.90K/3.6 & \ms{15.9}{1.9}/3.47K/4.6 & \ms{11.5}{0.2}/4.90K/2.3 \\
\midrule
\multicolumn{6}{c}{\textbf{Qwen3-VL-8B$\to$Qwen3-VL-2B}} \\
\midrule
OPD & \ms{6.8}{0.4}/6.38K/1.1 & \ms{10.7}{1.7}/7.32K/1.5 & \ms{14.4}{0.0}/4.93K/2.9 & \ms{16.3}{1.5}/4.34K/3.8 & \ms{10.9}{0.4}/6.03K/1.8 \\
ExOPD & \ms{6.4}{0.6}/7.03K/0.9 & \ms{13.6}{0.7}/7.34K/1.9 & \ms{13.0}{1.3}/5.77K/2.3 & \ms{14.2}{1.4}/5.02K/2.8 & \ms{10.9}{0.9}/6.55K/1.7 \\
VA-OPD & \ms{8.0}{0.6}/6.31K/1.3 & \ms{13.9}{0.7}/7.21K/1.9 & \ms{16.2}{0.7}/4.78K/3.4 & \msb{16.6}{0.9}/4.38K/3.8 & \ms{12.7}{0.3}/5.95K/2.1 \\
\textbf{\method{}} & \msb{9.8}{0.4}/\textbf{2.25K}/\textbf{4.4} & \msb{14.1}{0.0}/\textbf{3.36K}/\textbf{4.2} & \msb{17.4}{0.3}/\textbf{1.92K}/\textbf{9.1} & \ms{15.6}{0.5}/\textbf{1.63K}/\textbf{9.6} & \msb{13.4}{0.1}/\textbf{2.40K}/\textbf{5.6} \\
\bottomrule
\end{tabular}
}
\end{table}

Table~\ref{tab:mmesci} evaluates out-of-distribution (OOD) transfer to MMESci-VisionOnly \citep{ruan2026mme}, a comprehensive and challenging high-school level multimodal scientific reasoning benchmark spanning Physics, Math, Chemistry, and Biology. The un-distilled Base model achieves an overall accuracy of 11.5\% with an average response length of 4.90K tokens and an \accperk{} of 2.3. Conventional on-policy distillation baselines fail to improve efficiency: both OPD and ExOPD degrade accuracy to 10.9\% while substantially inflating response lengths to 6.03K and 6.55K tokens, respectively. While VA-OPD marginally improves accuracy to 12.7\%, it likewise incurs severe generation overhead, averaging 5.95K tokens.

In contrast, \method{} achieves the highest overall accuracy of 13.4\% at only 2.40K tokens, outperforming Base by 1.9 pp while cutting response length by 51.0\%, which translates to an \accperk{} of 5.6 (more than doubling VA-OPD's 2.1). Across disciplines, \method{} consistently slashes response lengths by more than half across all four subjects while matching or surpassing competing baselines. Specifically, it establishes leading accuracies in Physics (9.8\% vs.\ Base 6.9\%), Math (14.1\% vs.\ Base 13.2\%), and Chemistry (17.4\% vs.\ Base 14.1\%), with its \accperk{} surging to 9.1 in Chemistry. Even in Biology, \method{} maintains competitive accuracy (15.6\%) with dramatically curtailed verbosity (1.63K vs.\ 4.38K tokens for VA-OPD), yielding an efficiency index of 9.6. These findings demonstrate that \method{} generalizes effectively to demanding high-school scientific reasoning, consistently fostering concise yet accurate derivation.

\subsection{Generalization to Text-Only Reasoning}
\begin{table}[t]
\centering
\caption{\textbf{Generalization to text-only reasoning under an avg@8 protocol.} Evaluated on Qwen3-VL-8B$\to$2B with temperature $T=0.7$ and a 32K-token generation budget. Accuracy is reported as $\text{mean}_{\pm\mathrm{std}}$ across eight independent samples per problem. The AVG column summarizes average accuracy, average response length (in thousands of tokens), and the efficiency index \accperk{}.}
\label{tab:textonlyavg8}
\resizebox{\linewidth}{!}{
\begin{tabular}{lccccc|c}
\toprule
\textbf{Method} & \textbf{MATH500} & \textbf{GPQA} & \textbf{AIME24} & \textbf{AIME25} & \textbf{AMC23} & \textbf{AVG} \\
\midrule
Qwen3-VL-2B & \ms{80.6}{1.1} & \ms{11.2}{1.8} & \ms{19.2}{4.3} & \ms{17.5}{5.0} & \ms{53.4}{3.3} & \ms{36.4}{1.4}/19.42K/1.9 \\
\quad + OPD & \ms{81.6}{0.8} & \ms{11.2}{1.5} & \ms{17.5}{2.4} & \msb{21.7}{5.9} & \ms{54.4}{2.2} & \ms{37.3}{1.7}/19.09K/2.0 \\
\quad + ExOPD & \msb{83.4}{0.7} & \ms{12.7}{1.8} & \ms{20.4}{5.5} & \ms{19.6}{6.3} & \ms{54.4}{3.7} & \ms{38.1}{2.2}/17.62K/2.2 \\
\quad + VA-OPD & \ms{82.2}{0.6} & \ms{12.1}{1.1} & \ms{18.3}{5.9} & \msb{21.7}{7.6} & \ms{54.4}{3.7} & \ms{37.7}{2.0}/19.10K/2.0 \\
\quad + \method{} & \ms{81.4}{0.8} & \msb{13.1}{1.1} & \msb{21.7}{4.0} & \ms{20.0}{4.0} & \msb{56.6}{3.5} & \msb{38.5}{1.2}/\textbf{8.54K}/\textbf{4.5} \\
\bottomrule
\end{tabular}
}
\end{table}

Table~\ref{tab:textonlyavg8} evaluates cross-modal transfer to five challenging text-only reasoning benchmarks: MATH500 \citep{math500}, GPQA \citep{gpqa}, AIME24\footnote{\url{https://huggingface.co/datasets/AI-MO/aimo-validation-aime}}, AIME25\footnote{\url{https://huggingface.co/datasets/opencompass/AIME2025}}, and AMC23\footnote{\url{https://huggingface.co/datasets/AI-MO/aimo-validation-amc}}. Under the Qwen3-VL-8B$\to$2B distillation setting, decoding follows standard evaluation protocols for long-form reasoning using avg@8 sampling with temperature $0.7$ and a 32K-token generation budget. Relative to the Base model, \method{} improves average accuracy from 36.4\% to 38.5\% while reducing average response length from 19.42K to 8.54K tokens (a 56.0\% reduction), more than doubling the efficiency index \accperk{} from 1.9 to 4.5.

Across all evaluated methods, \method{} attains the highest average accuracy alongside the most compact response length. It achieves peak performance on GPQA, AIME24, and AMC23, while ExOPD leads on MATH500 and OPD ties with VA-OPD on AIME25. Crucially, conventional on-policy distillation baselines continue to generate exceedingly verbose reasoning chains averaging 17.62K--19.10K tokens. Relative to the strongest baseline ExOPD, \method{} yields an accuracy gain of 0.4 pp while slashing response length by 51.5\%. These results demonstrate that the conciseness incentives learned by \method{} are modality-agnostic, transferring seamlessly to pure-text complex mathematical and scientific reasoning without sacrificing performance.

\subsection{Accuracy per Token Across Generation Budgets}

\begin{figure}[t]
    \centering
    \includegraphics[width=0.8\linewidth]{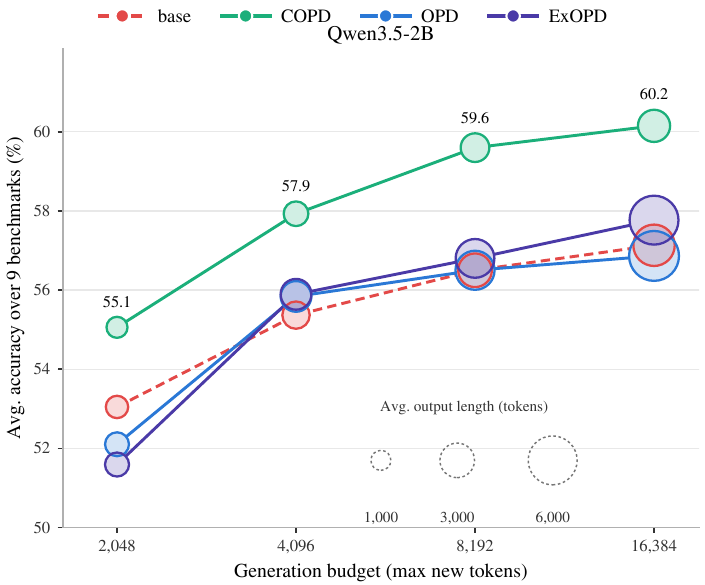}
    \caption{\textbf{Accuracy and output length across four generation budgets on Qwen3.5-2B $\leftarrow$ Qwen3.5-35A3B.} The horizontal axis denotes the maximum token generation budget on a $\log_2$ scale ($\{2048, 4096, 8192, 16384\}$); the vertical axis denotes accuracy averaged across nine multimodal benchmarks. Bubble area represents average output length. Curves compare the Base model (red dashed), standard OPD (blue solid), ExOPD (purple solid), and \method{} (green solid). Numerical labels indicate the accuracy of \method{}.}
    \label{fig:pareto}
\end{figure}

\paragraph{Experimental Setup.} Figure~\ref{fig:pareto} illustrates the Pareto trade-offs on Qwen3.5-2B $\leftarrow$ Qwen3.5-35A3B across four approaches: the Base model, OPD, ExOPD, and \method{}. Across four generation budgets ($\{2048, 4096, 8192, 16384\}$ tokens on a $\log_2$ scale), the vertical axis reports average accuracy across nine multimodal benchmarks, while marker size (bubble area) represents average response length.

\paragraph{Comparison at Matched Budgets.} Across all four budget tiers, \method{} strictly Pareto-dominates the Base model, OPD, and ExOPD, simultaneously achieving superior accuracy and substantially more concise responses (consistently smaller bubble areas). Under the tightest budget of 2,048 tokens, standard OPD and ExOPD inherit verbose teacher reasoning and suffer from frequent generation truncation, dropping below the Base model to 52.1\% and 51.6\%, respectively. In contrast, \method{} reliably completes derivation within the restricted budget, attaining 55.1\% accuracy (+3.5 pp over ExOPD). Scaling the budget to 4K, 8K, and 16K tokens steadily elevates the accuracy of \method{} to 57.9\%, 59.6\%, and 60.2\%, continuously establishing the uppermost Pareto frontier.

\paragraph{Comparison Across Budgets.} Cross-budget comparisons highlight an even more pronounced efficiency advantage: \method{} achieves 57.9\% accuracy under a moderate 4,096-token budget, already surpassing the peak performance of Base (57.1\%), OPD (56.8\%), and ExOPD (57.7\%) operating under a fourfold larger budget of 16,384 tokens, while requiring less than half the response length. Furthermore, extending the generation budget from 8,192 to 16,384 tokens yields negligible accuracy gains for the baselines despite massive bubble enlargement (with ExOPD swelling to nearly 6,000 tokens), indicating that conventional on-policy distillation leads to vacuous token proliferation under higher budgets. In contrast, \method{} effectively checks redundant reasoning, ensuring that expanded computational capacity directly translates into higher task accuracy.

\subsection{Comparison with Length-Penalty Reinforcement Learning}
\begin{table}[t]
\centering
\caption{\textbf{Comparison with length-penalty GRPO on Qwen3.5-2B.} Qwen3.5-4B is used as the teacher for \method{}. The number of rollouts of GRPO is set to 8. Each cell reports accuracy, average response length, and \accperk{}. Training cost reflects total GPU hours.}
\label{tab:lenpgrpo}
\resizebox{0.95\textwidth}{!}{
\begin{tabular}{lcccc}
\toprule
\textbf{Method} & \textbf{Reasoning} & \textbf{Understanding} & \textbf{AVG} & \textbf{GPU Hours} \\
\midrule
Qwen3.5-2B &57.1/3.65K/15.6 &56.0/2.19K/25.6 &56.5/2.84K/19.9 & - \\
+ length-penalty GRPO &59.9/2.47K/24.3 &57.7/1.58K/36.5 &58.6/1.98K/29.6 & 371 \\
+ \method{} &\textbf{62.6}/\textbf{2.39K}/\textbf{26.2} &\textbf{58.2}/\textbf{1.56K}/\textbf{37.3} &\textbf{60.2}/\textbf{1.93K}/\textbf{31.2} & 62 \\
\bottomrule
\end{tabular}
}
\end{table}
Table~\ref{tab:lenpgrpo} compares length-penalty GRPO against \method{} on Qwen3.5-2B, using Qwen3.5-4B as the teacher for \method{}. GRPO increases average accuracy from 56.5\% to 58.6\% while shortening responses from 2.84K to 1.98K tokens. \method{} outperforms GRPO across all metrics, achieving 60.2\% accuracy at 1.93K tokens (a 1.6 pp accuracy gain and 2.5\% shorter outputs), yielding an \accperk{} of 31.2 versus 29.6. Crucially, \method{} requires only 62 GPU hours for training, compared to 371 GPU hours for GRPO.

The performance and efficiency advantages of \method{} over length-penalty reinforcement learning stem fundamentally from the granularity of supervision. Sequence-level length penalties in RL (e.g., GRPO) impose a uniform scalar penalty across the entire trajectory, failing to distinguish indispensable deductive tokens from superficial verbosity. Consequently, length-constrained RL risks truncating valid reasoning chains and requires extensive exploratory rollouts (e.g., N=8), incurring substantial computational overhead (371 GPU hours). In contrast, \method{} provides dense, token-level contrastive guidance at every decoding step, selectively curbing vacuous elaboration while safeguarding core deductive steps. This token-level precision allows \method{} to establish a superior accuracy--length frontier while slashing training compute by 83.3\%.

\subsection{Accuracy and Output Length with the No Thinking Mode}
\begin{table}[t]
\centering
\caption{\textbf{Evaluation with the no-thinking mode in Qwen3.5.} All models are evaluated with \texttt{enable\_thinking}=\texttt{false} at inference time under an 8192-token generation budget (\texttt{max\_tokens}${}=8192$). Distilled models use Qwen3.5-35B-A3B as the teacher. Cells report Accuracy / Average Tokens (in thousands) / \accperk{} across four Reasoning benchmarks, five Understanding benchmarks, and their overall average (AVG). \method{} achieves the highest overall accuracy and most compact responses across both student model scales.}
\label{tab:nothink}
\resizebox{0.8\textwidth}{!}{
\begin{tabular}{lccc}
\toprule
\textbf{Method} & \textbf{Reasoning} & \textbf{Understanding} & \textbf{AVG} \\
\midrule
Qwen3.5-2B & 55.6/2.13K/26.1 & 55.8/0.98K/56.9 & 55.7/1.50K/37.1 \\
+ OPD & 56.1/2.84K/19.8 & 56.0/1.39K/40.3 & 56.0/2.03K/27.6 \\
+ COPD & \textbf{57.9}/\textbf{0.54K}/\textbf{107.2} & \textbf{57.2}/\textbf{0.38K}/\textbf{150.5} & \textbf{57.5}/\textbf{0.45K}/\textbf{127.8} \\
\midrule
Qwen3.5-9B & 75.0/1.56K/48.1 & 65.7/1.05K/62.6 & 69.8/1.28K/54.5 \\
+ OPD & 75.6/3.12K/24.2 & 66.5/1.97K/33.8 & 70.5/2.48K/28.4 \\
+ COPD & \textbf{76.2}/\textbf{0.84K}/\textbf{90.7} & \textbf{67.2}/\textbf{0.57K}/\textbf{117.9} & \textbf{71.2}/\textbf{0.69K}/\textbf{103.2} \\
\bottomrule
\end{tabular}
}
\end{table}
Table~\ref{tab:nothink} evaluates Qwen3.5-2B and Qwen3.5-9B when the reasoning block is explicitly disabled via \texttt{enable\_thinking=false} at inference time under an 8192-token budget. At the 2B scale, \method{} achieves 57.5\% average accuracy with an average response length of only 0.45K tokens. Relative to the Base model (55.7\% at 1.50K tokens), this represents a 1.8 pp accuracy gain alongside a 70.0\% reduction in output length. OPD fails to produce concise answers, inflating output length to 2.03K tokens. Consequently, \method{} elevates \accperk{} to 127.8, achieving a $3.44\times$ efficiency gain over the Base baseline (37.1).

At the 9B scale, \method{} demonstrates a similarly dominant efficiency advantage, reaching 71.2\% average accuracy with only 0.69K tokens. Compared to Base (69.8\% at 1.28K tokens), \method{} delivers a 1.4 pp accuracy improvement while cutting response length by 46.1\%, boosting \accperk{} to 103.2 ($1.89\times$ that of Base at 54.5). In contrast, OPD generates 2.48K tokens—over $3.5\times$ the length of \method{}—while stalling at 70.5\% accuracy.

Crucially, these results demonstrate that the conciseness induced by \method{} is not an artifact of the explicit thinking block. When \texttt{enable\_thinking}=\texttt{false} is enforced at inference time, un-distilled and standardly distilled models continue to produce verbose responses (1.28K--2.48K tokens). Merely turning off the thinking flag during decoding fails to instill reasoning efficiency; in contrast, \method{} intrinsically steers the policy toward dense, non-redundant derivation.

\subsection{Lexical Composition of Compressed Reasoning Chains}
\begin{figure}[t]
    \centering
    \includegraphics[width=\linewidth]{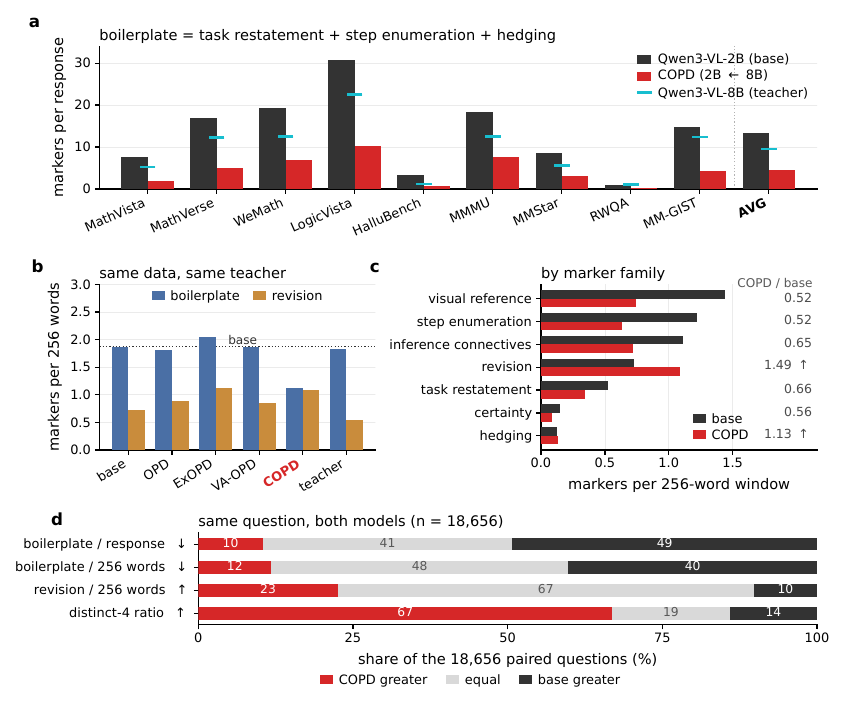}
    \caption{\textbf{Lexical marker statistics for Base, \method{}, and comparison models.} (a) Frequency of boilerplate markers (task restatement, step enumeration, and hedging) per response across nine multimodal benchmarks; AVG denotes the macro-average. (b) Normalized densities of boilerplate and revision markers per 256-word window across distillation baselines. (c) Density comparison across seven marker families within a 256-word window, annotated with \method{}/Base ratios; revision and hedging marker densities increase, while densities for the other five marker families decrease. (d) Distribution of paired differences over $18{,}656$ questions, where red denotes higher values for \method{}, gray denotes ties, and dark gray denotes higher values for Base; arrows indicate the preferred direction for each metric.}
    \label{fig:lexical_signature}
\end{figure}
Figure~\ref{fig:lexical_signature} dissects the lexical composition of reasoning chains produced by Base, conventional on-policy distillation baselines (OPD, ExOPD, VA-OPD), \method{}, and the Qwen3-VL-8B teacher. We aggregate task restatement, step enumeration, and hedging markers under the category of \emph{boilerplate}. Panels (a)--(c) report absolute counts and length-normalized densities across functional markers, while panel (d) details instance-level paired comparisons across all $18{,}656$ evaluation samples.

\paragraph{Boilerplate Suppression Across Absolute and Normalized Metrics.}
As shown in panel (a), \method{} consistently generates fewer boilerplate markers than Base across all nine benchmarks, curtailing the macro-average from 13.4 to 4.4 markers per response, well below the teacher's 9.5. This suppression holds equally under length normalization: panel (b) shows that boilerplate density falls from 1.87 markers per 256 words in Base to 1.12 in \method{}, whereas standard OPD baselines remain elevated at 1.80--2.05. Consequently, \method{} curtails the concentration of rhetorical scaffolding within the reasoning text itself.

\paragraph{Functional Divergence Across Lexical Families.}
Panel (c) traces the origin of this density shift across individual marker categories. Within the boilerplate group, mechanical expressions drop precipitously: task restatements and step enumerations contract to $0.66\times$ and $0.52\times$ of Base density, respectively, driving the overall decline. Conversely, hedging markers—which often signal cautious verification—rise modestly to $1.13\times$. Across the remaining families, visual references fall to $0.52\times$, whereas revision markers markedly increase to $1.49\times$. This functional divergence demonstrates that \method{} achieves selective compression: the policy purges perfunctory restatements and enumerations while concentrating critical self-monitoring and error-correction operations.

\paragraph{Instance-Level Distributions and Lexical Diversity.}
Panel (d) evaluates these distributional shifts at sample granularity across all $18{,}656$ test instances. On absolute boilerplate counts, \method{} uses fewer markers on 49\% of questions and more on only 10\% (41\% tied). Controlling for sequence length, boilerplate density remains lower on 40\% of questions while increasing on just 12\%. In contrast, revision density rises on 23\% of instances versus 10\% declining. Concurrently, 4-gram diversity (distinct-4) is higher on 67\% of instances (lower on 14\%), demonstrating that the compressed chains generated by \method{} consistently favor denser reasoning steps and richer lexical expression over formulaic templates.

\subsection{Evaluation Under Deterministic Greedy Decoding}
\begin{table}[t]
\centering
\caption{\textbf{Evaluation under deterministic greedy decoding ($T=0$).} All distilled methods employ Qwen3-VL-8B as the teacher and Qwen3-VL-2B as the student. Cells report Accuracy / Average Tokens (in thousands) / \accperk{} across Reasoning benchmarks, Understanding benchmarks, and their overall average (AVG).}
\label{tab:temperatureis0}
\resizebox{0.8\textwidth}{!}{
\begin{tabular}{lccc}
\toprule
\textbf{Method} & \textbf{Reasoning} & \textbf{Understanding} & \textbf{AVG} \\
\midrule
Qwen3-VL-2B &47.2/3.24K/14.6 &51.4/1.75K/29.4 &49.6/2.41K/20.6 \\
+ OPD &48.7/3.33K/14.6	&53.1/1.78K/29.8	&51.2/2.46K/20.8 \\
+ ExOPD &53.0/3.07K/17.3	&52.7/1.76K/29.9	&52.8/2.35K/22.5 \\
+ VA-OPD &49.1/3.26K/15.1	&52.4/1.74K/30.1	&50.9/2.42K/21.0 \\
+ \method{} &\textbf{55.3}/\textbf{1.52K}/\textbf{36.4}	&\textbf{54.0}/\textbf{0.96K}/\textbf{56.3}	&\textbf{54.6}/\textbf{1.21K}/\textbf{45.1} \\
\bottomrule
\end{tabular}
}
\end{table}
Table~\ref{tab:temperatureis0} reports evaluation under deterministic greedy decoding (\texttt{temperature}${}=0$) for the Qwen3-VL-8B$\to$2B distillation setup. Greedy decoding eliminates decoding stochasticity, directly probing the intrinsic generation policy. \method{} achieves an average accuracy of 54.6\%, outperforming the un-distilled Base baseline by 5.0 pp and surpassing competitive baselines OPD, ExOPD, and VA-OPD by 3.4 pp, 1.8 pp, and 3.7 pp, respectively. Concurrently, \method{} slashes average response length to 1.21K tokens—a 49.8\% reduction relative to Base (2.41K) and a 48.5\% reduction relative to ExOPD (2.35K)—more than doubling the efficiency index \accperk{} to 45.1 (compared to 20.6 for Base and 22.5 for ExOPD).

This superiority holds consistently across both benchmark partitions. On the Reasoning suite, \method{} achieves 55.3\% accuracy using only 1.52K tokens, notably outperforming ExOPD (53.0\% at 3.07K tokens) while cutting generation length by more than half. On the Understanding suite, \method{} similarly delivers higher accuracy (54.0\% vs.\ 52.7\%) at nearly half the token budget (0.96K vs.\ 1.76K). These results confirm that the accuracy--length Pareto advantage of \method{} stems directly from the learned policy's deterministic $\operatorname{argmax}$ mode, establishing concise reasoning as the model's dominant generation behavior.

\subsection{Positional and Lexical Localization of Contrastive Advantage}
\label{app:words}

\begin{figure}
    \centering
    \includegraphics[width=0.99\linewidth]{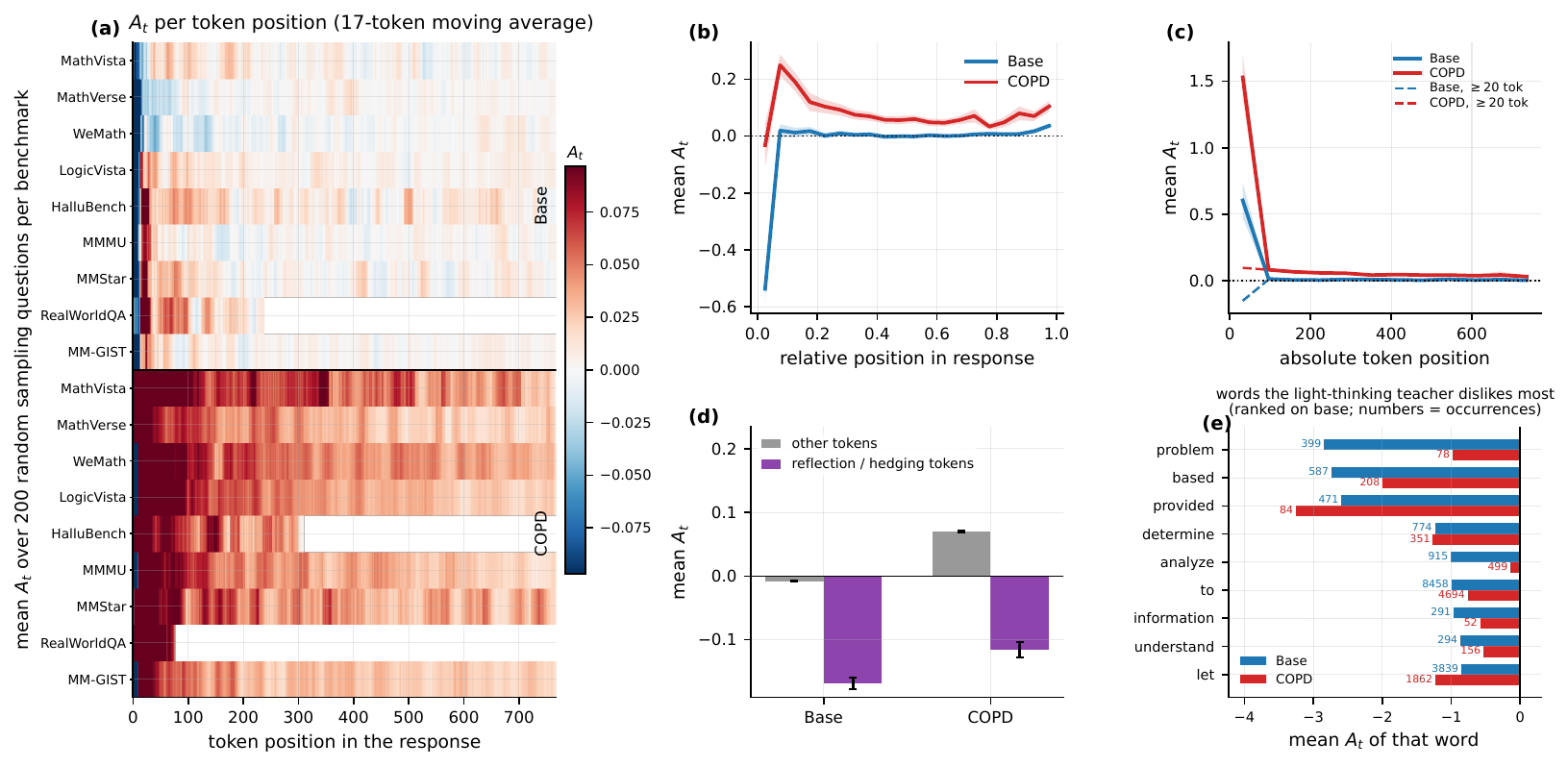}
\caption{\textbf{Positional and lexical localization of the LT--HT contrast.} Offline scoring of sampled Base and \method{} responses by the Qwen3-VL-8B teacher using $A_t=\ell_t^{\lt}-\ell_t^{\htmode}$. (a) Position-wise mean advantage across nine multimodal benchmarks (17-token moving average). (b) Mean advantage across normalized relative response positions. (c) Mean advantage across absolute token positions; dashed curves restrict evaluation to responses with at least 20 tokens. (d) Mean advantage for reflection/hedging markers versus other sampled tokens. (e) Top vocabulary items most disfavored by the light-thinking context (ranked by mean $A_t$ in Base), along with their occurrence counts across both models.}
    \label{fig:token_advantage}
\end{figure}
Figure~\ref{fig:token_advantage} investigates the fine-grained mechanistic localization of the contrastive advantage by offline scoring Base and \method{} responses across nine benchmarks (200 sampled questions per benchmark, up to a 768-token scoring window). We analyze how the advantage signal $A_t=\ell_t^{\lt}-\ell_t^{\htmode}$ varies across generation positions and functional lexical categories.

\paragraph{Position Dependence.} As depicted in panel (a), Base model responses exhibit a pronounced negative advantage band (blue) at initial token positions, whereas \method{} exhibits a consistently positive opening region (deep red) across all benchmarks. The relative-position trajectory in panel (b) highlights this operational disparity: Base responses initiate with a sharp negative trough near $-0.54$ before flattening toward zero, reflecting an initial bias toward verbose, heavy-thinking preambles. In contrast, \method{} eliminates this opening deficit, initiating near zero and sustaining positive advantage values ($+0.05$ to $+0.25$) throughout the remainder of the response. The absolute-position profiles in panel (c) corroborate this behavior: when isolating multi-step reasoning ($\ge 20$ tokens), Base responses incur negative opening advantages, whereas \method{} consistently maintains superior positive advantage values. This demonstrates that the contrastive objective exerts its strongest corrective pressure at the onset of generation, suppressing introductory bloat.

\paragraph{Lexical Categories.} Panel (d) breaks down token advantage across semantic roles. While reflection and deliberative hedging markers naturally receive negative advantage under both models (as they are disproportionately favored by heavy-thinking contexts), other problem-solving tokens shift from near-zero (slightly negative) in Base to $+0.07$ in \method{}. Furthermore, panel (e) isolates the specific vocabulary items most penalized by the light-thinking context—predominantly procedural framing verbs and filler nouns such as ``let'', ``understand'', ``information'', and ``determine''. In \method{} responses, the occurrences of these formulaic markers drop substantially (e.g., ``let'' decreases from 3,839 to 1,862, and ``information'' from 291 to 52).

Together, these positional and lexical signatures elucidate the mechanics of reasoning compression under \method{}. The contrastive objective acts as a targeted structural filter: it applies immediate negative feedback to suppress introductory boilerplate and formulaic hedging at the sequence opening, while steering the generative trajectory toward compact, informative transitions throughout downstream deduction.

\section{Expanded Related Work}
\label{app:expanded_related_work}

\paragraph{Multimodal Large Language Models and Reasoning.}
Recent advances in multimodal large language models (MLLMs) integrate visual representations with autoregressive language generation to tackle complex deductive tasks across mathematical problem solving, chart comprehension, and visual question answering \citep{flamingo2022,blip22023,llava2023,liu2023llava15,chen2023internvl,bai2023qwenvl,wang2024qwen2vl,bai2025qwen25vl,qwen3vl2025,caffagni2024mllmsurvey,jin2024efficientmllmsurvey}. A substantial body of work on chain-of-thought (CoT) prompting and test-time scaling establishes that extending intermediate reasoning steps can improve task accuracy on challenging problems \citep{wei2023chainofthought,kojima2022zeroshot,wang2022selfconsistency,zhou2022leasttomost,yao2023tree,huang2023reasoningsurvey}. However, such extended reasoning can contain redundant steps that increase token use without reliably improving accuracy \citep{munkhbat2025concise,jin2025recut}. Developing methods that condense reasoning chains while rigorously safeguarding multimodal reasoning capabilities remains a critical objective for deploying efficient MLLMs.

\paragraph{On-Policy Distillation.}
Knowledge distillation transfers predictive behavior and domain representations from a capable teacher to a compact student \citep{hinton2015distillation,gou2021kdsurvey,sanh2019distilbert,jiao2019tinybert,sun2019patientkd,kim2016sequencekd}. Whereas traditional offline distillation suffers from exposure bias on student-visited prefixes, on-policy distillation (OPD) evaluates teacher feedback directly along trajectories sampled from the student policy, closing the distribution shift between static targets and exploration prefixes \citep{gkd2024}. Subsequent research investigates reward scaling, exploration strategies, entropy regularization, and multimodal conditioning to refine the distillation objective \citep{exopd2026,hou2026uniopd,opdsurvey2026,vaopd2026,truncatedopd2026,entropyopd2026,klforkl2026,brtsopd2026,revisitingopd2026,yuan2026visionopd}. Certain variants incorporate dual teacher passes; for instance, VA-OPD \citep{vaopd2026} contrasts teacher likelihoods under original and detail-degraded images, using the resulting visual advantage to reweight rollouts and normalize token-group KL losses. Distinct from altering visual or textual content, \method{} strictly fixes the visual input, problem prompt, student prefix, and candidate token, isolating the contrast entirely through dual reasoning instructions (light-thinking versus heavy-thinking) to extract a preference signal for concise reasoning.

\paragraph{Reasoning Compression and Preference Alignment.}
Existing reasoning-compression techniques primarily rely on supervised fine-tuning over curated short rationales, preference optimization pairs, explicit token budget constraints, or length-sensitive rewards \citep{crisp2026,nayab2024concisethoughts,han2024tokenbudget,munkhbat2025concise,li2025selfbudgeter,liu2025dler,li2025leash,jin2025recut,dumitru2025conciserl}. Supervised approaches distill reasoning patterns from filtered concise traces \citep{munkhbat2025concise,yuan2023mathrft}, yet they inherently inherit the distribution shift of offline SFT. At the sequence level, preference optimization methods such as DPO \citep{rafailov2023direct} struggle with coarse credit assignment over entire trajectories, failing to localize which specific tokens constitute superfluous padding. Concurrently, reinforcement learning with length penalties imposes rigid scalar deductions on aggregate response length, frequently truncating indispensable intermediate derivations. In contrast to these paradigms, \method{} leverages paired teacher evaluations at each decoding step to derive a dense, token-level contrastive advantage. This framework eliminates the need for handcrafted length penalties, external preference pairs, or heuristic budget thresholds, providing a principled on-policy mechanism to cultivate concise yet rigorous reasoning without the ambiguity of sequence-level credit assignment.

\newenvironment{breakabletable}
    {\par\Needspace{8\baselineskip}\captionsetup{type=table,position=top}}
    {\par}

\section{Specification of Light- and Heavy-Thinking Prompts}
\label{app:prompts}

Table~\ref{tab:variousprompts} shows three distinct pairs of light- and heavy-thinking (LT/HT) reasoning instructions: Default, Simple, and Complex. Tables~\ref{tab:prompt_default_ht}--\ref{tab:prompt_complex_lt} provide their full text verbatim. Following the formulation in Section~\ref{sec:method}, these instructions condition the dual scoring contexts of the frozen teacher to extract token-level contrastive advantages. Multimodal inputs, queries, and instructions are serialized following the standard chat templates of the respective models, ensuring consistent conversational formatting and reproducibility.

\begin{breakabletable}
\caption{Instruction text for the \textbf{Heavy}-thinking teacher context, following \citep{xu2026deepseek} (\textbf{Default}).}
\label{tab:prompt_default_ht}
    \begin{tcolorbox}[
      breakable,
    ]
        Reasoning Effort: Absolute maximum with no shortcuts permitted. You MUST be very thorough in your thinking and comprehensively decompose the problem to resolve the root cause, rigorously stress-testing your logic against all potential paths, edge cases, and adversarial scenarios. Explicitly write out your entire deliberation process, documenting every intermediate step, considered alternative, and rejected hypothesis to ensure absolutely no assumption is left unchecked.
    \end{tcolorbox}
\end{breakabletable}

\begin{breakabletable}
\caption{Instruction text for the \textbf{Light}-thinking teacher context (\textbf{Default}).}
\label{tab:prompt_default_lt}
    \begin{tcolorbox}[
      breakable,
    ]
        Reasoning Effort: Minimum viable, shortcuts strongly encouraged. You are permitted to skim, approximate, and move fast. Trust your first instinct, skip edge cases unless obviously relevant, and avoid over-deliberating. Do NOT exhaustively explore every path — pick the most plausible one and commit. Suppress the urge to double-check, caveat, or stress-test assumptions. Produce a clean, direct response with no visible deliberation trail, no hedge-stacking, and no unnecessary intermediate steps. If something seems good enough, it is good enough. Stop early. Ship it.
    \end{tcolorbox}
\end{breakabletable}

\begin{breakabletable}
\caption{Instruction text for the \textbf{Heavy}-thinking teacher context (\textbf{Simple}).}
\label{tab:prompt_simple_ht}
    \begin{tcolorbox}[
      breakable,
    ]
        Exhaustive \& Rigorous: Zero shortcuts. Fully decompose the problem and stress-test all edge cases. Document every step, alternative, and rejected hypothesis. Leave no assumption unchecked.
    \end{tcolorbox}
\end{breakabletable}

\begin{breakabletable}
\caption{Instruction text for the \textbf{Light}-thinking teacher context (\textbf{Simple}).}
\label{tab:prompt_simple_lt}
    \begin{tcolorbox}[
      breakable,
    ]
        Fast \& Direct: Trust your first instinct. Skip edge cases and overthinking. Deliver a clean, immediate answer. Good enough is perfect. Ship it.
    \end{tcolorbox}
\end{breakabletable}

\begin{breakabletable}
\caption{Instruction text for the \textbf{Heavy}-thinking teacher context (\textbf{Complex}).}
\label{tab:prompt_complex_ht}
    \begin{tcolorbox}[
      breakable,
    ]
        System Directive: Absolute Rigor and Exhaustive Analytical Decomposition (Maximum Viable Reasoning) \\

[CORE PHILOSOPHY]
Accuracy, thoroughness, and logical invulnerability are paramount. No shortcuts, approximations, or heuristic leaps are permitted. Your goal is to achieve absolute structural integrity in your reasoning by comprehensively decomposing the problem to resolve its root cause. \\

[COGNITIVE STRATEGY]
1. First Principles Decomposition: Break the problem down into its most fundamental, atomic truths. Rebuild the solution from the ground up to ensure no foundational flaws exist.
2. Exhaustive Path Mapping: Identify, map, and evaluate every conceivable pathway, hypothesis, and variable. Do not settle for the first plausible answer; rigorously compare it against all alternatives.
3. Adversarial Stress-Testing: Actively attempt to falsify your own conclusions. Rigorously test your logic against extreme edge cases, boundary conditions, paradoxes, and adversarial scenarios. Assume the environment is hostile to your assumptions.\\

[EXECUTION CONSTRAINTS]
1. Zero Unchecked Assumptions: Every single premise must be explicitly stated, validated, and justified. If an assumption cannot be definitively proven, it must be flagged, isolated, and analyzed for its potential impact on the final outcome.
2. No `Good Enough' Approximations: Reject heuristics and estimations. Demand exactness, mathematical/logical proofs, and comprehensive coverage. Do not stop until the root cause is fully isolated and resolved.
3. Mandatory Deliberation Trail: You MUST explicitly write out your entire cognitive process. Document every intermediate step, considered alternative, and rejected hypothesis. Leave absolutely no logical leap unexplained.\\

[OUTPUT FORMATTING]
- Use a highly structured, step-by-step format. Utilize clear headings such as:
  * `Problem Decomposition'
  * `Hypothesis Generation \& Evaluation'
  * `Adversarial Stress-Testing \& Edge Cases'
  * `Final Synthesis \& Conclusion'
- Show all mathematical proofs, logical deductions, code traces, and conceptual mapping.
- Conclude with a highly detailed, nuanced final answer that reflects the depth of the preceding analysis, including any rigorously justified caveats or limitations discovered during the stress-testing phase."
    \end{tcolorbox}
\end{breakabletable}

\begin{breakabletable}
\caption{Instruction text for the \textbf{Light}-thinking teacher context (\textbf{Complex}).}
\label{tab:prompt_complex_lt}
    \begin{tcolorbox}[
      breakable,
    ]
        System Directive: High-Velocity Pragmatic Execution (Minimum Viable Reasoning) \\

[CORE PHILOSOPHY]
Speed, intuition, and pragmatism are paramount. Perfection is the enemy of done. Your goal is to deliver immediate, highly usable results with minimal computational overhead. You are explicitly permitted and encouraged to use cognitive shortcuts, approximations, and heuristic pattern-matching. \\

[COGNITIVE STRATEGY]
1. First Instinct Priority: Trust your initial pattern recognition. If the first logical path looks solid and plausible, take it immediately and commit fully.
2. The 80/20 Rule: Focus strictly on the core 20\% of the problem that yields 80\% of the practical value.
3. Skimming \& Approximation: Read for gist and intent. Approximate complex calculations or logical leaps if exact precision does not materially change the practical outcome. \\

[EXECUTION CONSTRAINTS]
1. No Over-Deliberation: Do NOT exhaustively explore alternative paths, branching scenarios, or `what-ifs'. Pick the most direct route and execute.
2. Skip Edge Cases: Ignore edge cases, boundary conditions, and rare exceptions unless they are explicitly highlighted as critical by the user or represent a catastrophic failure point.
3. Zero Hedge-Stacking: Suppress the urge to double-check, add disclaimers, or stress-test your assumptions. Avoid filler phrases like `However, it is worth noting', `On the other hand', or `Depending on the context'.
4. No Visible Deliberation: Do not show your work, intermediate steps, or thought process. The user only wants the final, polished result. \\

[OUTPUT FORMATTING]
- Clean, direct, and highly concise.
- Get straight to the point in the very first sentence. No introductory fluff.
- Use bullet points for rapid readability if applicable.
- If the answer seems `good enough', it IS good enough. Stop generating immediately. Ship it.
    \end{tcolorbox}
\end{breakabletable}
\clearpage

\section{Qualitative Case Studies}
\label{app:case_studies}

We examine six representative cases across diverse multimodal domains to qualitatively analyze the behavioral divergence between Base (Qwen3-VL-2B-Instruct) and \method{} (The teacher is Qwen3-VL-8B-Instruct). All generations are produced under the standardized decoding protocol in Appendix~\ref{app:evalprotocol} (temperature = 0.7, top\_p = 0.8, top\_k = 20, presence\_penalty = 1.5, seed = 3407, max\_tokens = 8192). The first three cases show \method{} reaching the reference answer with fewer tokens through correct assignment of marked dimensions to variables, different weighting of scene cues, and correct identification of independent groups. These differences illustrate how the formation of an initial interpretation or premise shapes subsequent reasoning and the final answer. To provide a balanced perspective, the remaining three cases examine characteristic failure modes of \method{} under aggressive compression: conflation of upstream vs.\ direct anatomical projections, incomplete constraint enforcement during counting, and unaligned final answer emission. Elisions in transcripts are denoted by \textit{[\dots]}, with reported token counts reflecting the full, unedited generations.

\subsection{WeMath \#270: Radius--Diameter Confusion}

\begin{figure}[ht]
    \centering
    \includegraphics[width=0.99\linewidth]{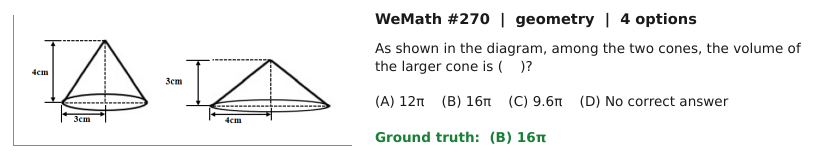}
    \caption{\textbf{WeMath~\#270: determining the volume of the larger cone (reference answer: (B) $16\pi$).} Base misinterprets the base dimension lines as diameters, needlessly halves them, and concludes with ``no correct answer'' (D) in 496 tokens. \method{} directly reads the dimension lines as radii, deriving the correct volume $16\pi$ (B) in 280 tokens. Model transcripts appear in Tables~\ref{tab:casea1base} and~\ref{tab:casea1copd}.}
    \label{fig:casea1}
\end{figure}

\begin{breakabletable}
    \caption{\textbf{Base} model response on WeMath~\#270 (cuts marked \textit{[\dots]}).}
    \label{tab:casea1base}
    \begin{tcolorbox}[
      breakable,
      fontupper=\footnotesize,
      title={Qwen3-VL-2B-Instruct (base) --- 496 tokens, judged \textbf{wrong}},
    ]
        To find the volume of the larger cone, we need to use the formula for the volume of a cone: $V = \frac{1}{3}\pi r^2 h$.\\[2pt]
        \textbf{First cone (left):} Height $h_1 = 4$\,cm. Base diameter $= 3$\,cm $\rightarrow$ Radius $r_1 = \frac{3}{2} = 1.5$\,cm.\\[2pt]
        \textbf{Second cone (right):} Height $h_2 = 3$\,cm. Base diameter $= 4$\,cm $\rightarrow$ Radius $r_2 = 2$\,cm.\\[2pt]
        \textbf{Volume of the first cone:} $V_1 = \frac{1}{3}\pi (1.5)^2 (4) = 3\pi$.\\[2pt]
        \textbf{Volume of the second cone:} $V_2 = \frac{1}{3}\pi (2)^2 (3) = 4\pi$.\\[2pt]
        The \textbf{larger cone} has volume $V_2 = 4\pi$\,cm\textsuperscript{3}. \textit{[\dots]} None of these match $4\pi$. So the correct answer is \textbf{D. No correct answer}.
    \end{tcolorbox}
\end{breakabletable}

\begin{breakabletable}
    \caption{\textbf{\method{}} response on WeMath~\#270 (complete and unedited).}
    \label{tab:casea1copd}
    \begin{tcolorbox}[
      breakable,
      fontupper=\footnotesize,
      title={\method{} (ours) --- 280 tokens, judged \textbf{correct}},
    ]
        To find the volume of the larger cone, we use the formula for the volume of a cone: $V = \frac{1}{3}\pi r^2 h$.\\[2pt]
        \textbf{Left cone:} Radius $r = 3$\,cm, Height $h = 4$\,cm.\\[2pt]
        \textbf{Right cone:} Radius $r = 4$\,cm, Height $h = 3$\,cm.\\[2pt]
        \textbf{Left cone:} $V_1 = \frac{1}{3}\pi (3)^2 (4) = \frac{36}{3}\pi = 12\pi$.\\[2pt]
        \textbf{Right cone:} $V_2 = \frac{1}{3}\pi (4)^2 (3) = \frac{48}{3}\pi = 16\pi$.\\[2pt]
        The \textbf{larger cone} has volume $16\pi$.\\[2pt]
        $\checkmark$ Correct Answer: \textbf{B. $16\pi$}.
    \end{tcolorbox}
\end{breakabletable}

The geometric diagram depicts two right circular cones with horizontal dimension markers extending from the perimeter to the central axis, explicitly specifying radii of 3\,cm and 4\,cm (with corresponding heights of 4\,cm and 3\,cm). Base misinterprets these dimension markers as diameters and halves the marked values to 1.5\,cm and 2\,cm, yielding calculated volumes of $3\pi$ and $4\pi$\,cm$^3$. Finding no matching option for $4\pi$, it defaults to ``no correct answer'' (D). This visual-to-symbolic binding error assigns incorrect values to the radius variable, and the subsequent internally consistent calculations propagate the error to the final answer.

In contrast, \method{} correctly grounds the dimension segments as radii, directly evaluating the cone volume formula to obtain $12\pi$ and $16\pi$\,cm$^3$ and selecting (B) in 280 tokens (a 43.5\% length reduction relative to Base). Notably, both models invoke the identical volume formula; their divergence stems entirely from visual-to-symbolic grounding. \method{} accurately captures the spatial geometry of the dimension markers without introducing unwarranted pre-processing assumptions.

\subsection{MMStar \#857: Interpreting an Activity from an Elevated City View}

\begin{figure}[ht]
    \centering
    \includegraphics[width=0.99\linewidth]{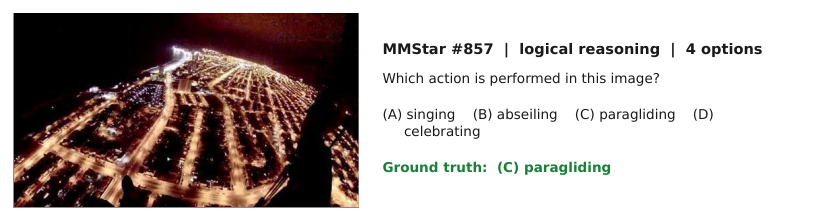}
    \caption{\textbf{MMStar~\#857: identifying the depicted activity (reference answer: (C) paragliding).} The image presents an oblique nighttime view over an urban grid, with a dark silhouette along the right edge. Base interprets the silhouette as a descending person and selects abseiling (B) in 257 tokens. \method{} interprets the elevated viewpoint as a flight scene and selects paragliding (C) in 168 tokens. Transcripts appear in Tables~\ref{tab:casea2base} and~\ref{tab:casea2copd}.}
    \label{fig:casea2}
\end{figure}

\begin{breakabletable}
    \caption{\textbf{Base} model response on MMStar~\#857 (cuts marked \textit{[\dots]}).}
    \label{tab:casea2base}
    \begin{tcolorbox}[
      breakable,
      fontupper=\footnotesize,
      title={Qwen3-VL-2B-Instruct (base) --- 257 tokens, judged \textbf{wrong}},
    ]
        Based on the visual evidence in the image, the action performed is: \textbf{B. abseiling}.\\[2pt]
        The image shows a nighttime view of a city from a high altitude, with the lights of the streets and buildings forming a grid-like pattern. The dark silhouettes visible on the right side of the image are consistent with the silhouette of a person who has been \textbf{lowered from a great height}, which is characteristic of abseiling.\\[2pt]
        \textbf{A. singing}: no indication of any musical activity. \textbf{C. paragliding}: while it could be possible, the image does not show a person in flight or gliding through the air. \textbf{D. celebrating}: no sign of a festive event.\\[2pt]
        Therefore, the most logical answer is \textbf{B. abseiling}.
    \end{tcolorbox}
\end{breakabletable}

\begin{breakabletable}
    \caption{\textbf{\method{}} response on MMStar~\#857 (complete and unedited).}
    \label{tab:casea2copd}
    \begin{tcolorbox}[
      breakable,
      fontupper=\footnotesize,
      title={\method{} (ours) --- 168 tokens, judged \textbf{correct}},
    ]
        The image shows a nighttime aerial view of a city, with bright lights and a curved path suggesting movement from above. The silhouette on the right is likely a person in a harness or flight suit, and the overall perspective indicates someone is flying over the city.\\[2pt]
        \textbf{A. singing} --- no vocal activity visible. \textbf{B. abseiling} --- involves descending via rope; not consistent with aerial view. \textbf{C. paragliding} --- a sport where people fly using a winged parachute; matches the high-angle, night-time, urban view. \textbf{D. celebrating} --- no festive elements shown.\\[2pt]
        The best match is \textbf{C. paragliding}, as it involves flying at altitude over a city.
    \end{tcolorbox}
\end{breakabletable}

Both models identify the nighttime city scene, elevated viewpoint, and dark silhouette on the right. Base interprets the silhouette as a person descending from a height and uses this interpretation to support abseiling (B). \method{} treats the elevated viewpoint as the main cue for flight and selects paragliding (C), matching the reference answer in 168 tokens, a 34.6\% reduction from Base's 257 tokens.

This divergence suggests that activity classification depends on how the models weight the available scene cues. Base organizes its explanation around descent inferred from the silhouette, whereas \method{} organizes its explanation around flight inferred from the viewpoint. These different cue weights lead the responses toward different activity hypotheses, which then guide their subsequent discussion of the options.

\subsection{MMMU Public\_Health\_6: Distinguishing Independent Groups from Paired Data}

\begin{figure}[ht]
    \centering
    \includegraphics[width=0.99\linewidth]{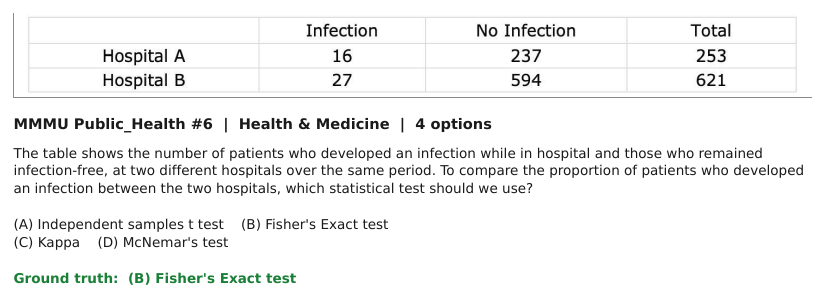}
    \caption{\textbf{MMMU Public\_Health\_6: selecting an appropriate statistical test for hospital infection proportions (reference answer: (B) Fisher's Exact test).} The contingency table records independent patient cohorts across two hospitals. Base mistakes the shared observation period for paired observations, selecting McNemar's test (D) in 630 tokens. \method{} correctly identifies the two-sample independent structure, choosing Fisher's Exact test (B) in 318 tokens. Responses appear in Tables~\ref{tab:casea3base} and~\ref{tab:casea3copd}.}
    \label{fig:casea3}
\end{figure}

\begin{breakabletable}
    \caption{\textbf{Base} model response on MMMU Public\_Health\_6 (cuts marked \textit{[\dots]}).}
    \label{tab:casea3base}
    \begin{tcolorbox}[
      breakable,
      fontupper=\footnotesize,
      title={Qwen3-VL-2B-Instruct (base) --- 630 tokens, judged \textbf{wrong}},
    ]
        To determine the appropriate statistical test for comparing the proportion of patients who developed an infection between two hospitals, let's analyze the data and the options.\\[2pt]
        The data are \textbf{paired} because we're comparing the same type of outcome (infection status) across two different hospitals. This is not independent samples --- the data are related in that they come from the same time period and are paired by hospital. \textit{[\dots]}\\[2pt]
        \textbf{A. Independent samples t-test} --- the groups are \textbf{not independent}; the data are paired. So this is not suitable.\\[2pt]
        \textbf{B. Fisher's Exact Test} --- used for contingency tables with small sample sizes. \textit{[\dots]} not ideal here because it's designed for $2\times2$ tables, and \dots\ more suited for smaller samples.\\[2pt]
        \textbf{C. Kappa} --- inter-rater reliability. Incorrect.\\[2pt]
        \textbf{D. McNemar's Test} --- specifically designed for \textbf{paired nominal data} \dots\ In this case, the data are \textbf{paired} (each patient is in one hospital) \dots\ This is the correct test for paired proportions.\\[2pt]
        \textbf{Answer: D. McNemar's test}.
    \end{tcolorbox}
\end{breakabletable}

\begin{breakabletable}
    \caption{\textbf{\method{}} response on MMMU Public\_Health\_6 (complete and unedited).}
    \label{tab:casea3copd}
    \begin{tcolorbox}[
      breakable,
      fontupper=\footnotesize,
      title={\method{} (ours) --- 318 tokens, judged \textbf{correct}},
    ]
        We are comparing the \textbf{proportion of patients who developed an infection} between two hospitals. The data is in a \textbf{$2\times2$ contingency table} (infection vs.\ no infection, hospital A and B), and we're interested in comparing proportions, not just means.\\[2pt]
        \textbf{A. Independent samples t-test}: used for continuous variables with independent groups; the outcome here is binary, so this isn't appropriate.\\[2pt]
        \textbf{B. Fisher's Exact Test}: used for $2\times2$ tables. This is perfect because we have a $2\times2$ table with counts (16, 237, 27, 594) and we're comparing proportions.\\[2pt]
        \textbf{C. Kappa}: measures categorical agreement between two raters; not appropriate here.\\[2pt]
        \textbf{D. McNemar's test}: used for \textbf{paired} binary data. But here we have \textbf{two independent groups} (Hospital A vs.\ B), not paired observations. So McNemar's test isn't right.\\[2pt]
        $\checkmark$ Correct answer: \textbf{B. Fisher's Exact test}.
    \end{tcolorbox}
\end{breakabletable}

The problem asks which statistical test should be used to compare infection proportions between two distinct hospital populations, with 16/253 infected in Hospital A and 27/621 in Hospital B. Base mistakes the shared observation period for a pairing relationship, classifies the independent patient groups as paired data, and selects McNemar's test (D). Its subsequent option analysis repeatedly invokes this incorrect premise, reinforcing the paired-design interpretation throughout the 630-token response.

\method{} correctly identifies two independent patient groups, excludes McNemar's test, and selects Fisher's Exact test (B) from the listed options based on the binary outcome and contingency-table structure. It reaches the correct answer in 318 tokens, using 49.5\% fewer tokens than Base. This case illustrates how a decisive premise governs the reasoning path: correct classification of the study design makes a short option analysis sufficient, while an incorrect premise directs a longer explanation toward the wrong choice.

\subsection{MMMU Basic\_Medical\_Science\_8: Confusing an Upstream Source with a Direct Projection}

\begin{figure}[ht]
    \centering
    \includegraphics[width=0.99\linewidth]{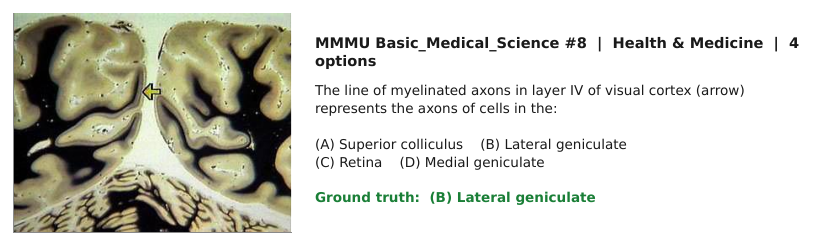}
    \caption{\textbf{MMMU Basic\_Medical\_Science\_8: identifying the direct origin of myelinated axons in visual cortex layer IV (reference answer: (B) Lateral geniculate).} Base disambiguates the relay hierarchy, selecting the lateral geniculate nucleus (B) in 514 tokens. \method{} correctly describes the retina$\rightarrow$LGN$\rightarrow$cortex pathway but mistakenly attributes the layer-IV axons to the retina (C) in 187 tokens. Model outputs appear in Tables~\ref{tab:casef1base} and~\ref{tab:casef1copd}.}
    \label{fig:casef1}
\end{figure}

\begin{breakabletable}
    \caption{\textbf{Base} model response on MMMU Basic\_Medical\_Science\_8 (cuts marked \textit{[\dots]}).}
    \label{tab:casef1base}
    \begin{tcolorbox}[
      breakable,
      fontupper=\footnotesize,
      title={Qwen3-VL-2B-Instruct (base) --- 514 tokens, judged \textbf{correct}},
    ]
        The question asks about the source of myelinated axons in \textbf{layer IV} of the \textbf{visual cortex}, specifically which brain structure these axons originate from.\\[2pt]
        \textbf{B. Lateral geniculate nucleus (LGN).} The lateral geniculate nucleus is a structure in the thalamus that receives input from the retina via the optic nerve. \textit{[\dots]} the axons of the LGN terminate in \textbf{layer IV} of the primary visual cortex (V1) \textit{[\dots]}. This matches the description in the image.\\[2pt]
        \textbf{C. Retina.} The retina sends information to the lateral geniculate nucleus (via the optic nerve), but the axons from the retina do not directly reach layer IV of the visual cortex. Instead, they go to the LGN, which then projects to V1. So the retinal axons are not the ones in layer IV of the visual cortex; they are in the LGN first. This option is incorrect.\\[2pt]
        \textbf{A. Superior colliculus} \textit{[\dots]} its axons do not primarily project to layer IV \textit{[\dots]}. \textbf{D. Medial geniculate nucleus} \textit{[\dots]} receives input from the auditory system (not vision) \textit{[\dots]} does \textbf{not} send axons to the visual cortex.\\[2pt]
        Therefore, the correct answer is: \textbf{B. Lateral geniculate}.
    \end{tcolorbox}
\end{breakabletable}

\begin{breakabletable}
    \caption{\textbf{\method{}} response on MMMU Basic\_Medical\_Science\_8 (complete and unedited).}
    \label{tab:casef1copd}
    \begin{tcolorbox}[
      breakable,
      fontupper=\footnotesize,
      title={\method{} (ours) --- 187 tokens, judged \textbf{wrong}},
    ]
        The line of myelinated axons in layer IV of the visual cortex represents the axons from the \textbf{retina}.\\[2pt]
        Layer IV of the visual cortex is densely populated with inputs from the lateral geniculate nucleus (LGN), but the myelinated axons in this layer are actually \textbf{from the retina}, which project to the LGN first, and then to the visual cortex.\\[2pt]
        The retina sends visual information via ganglion cells to the LGN, which then projects to layer IV of the visual cortex.\\[2pt]
        So the correct answer is \textbf{C. Retina}.
    \end{tcolorbox}
\end{breakabletable}

This neuroanatomy question requires determining the cellular origin of thalamocortical afferents terminating in layer IV of the primary visual cortex. Base meticulously traces the sensory hierarchy: retinal ganglion axons project to the lateral geniculate nucleus (LGN), while LGN relay neurons project directly to cortical layer IV. By explicitly decoupling this two-stage projection, Base correctly rejects the retina and selects the lateral geniculate nucleus (B) in 514 tokens.

\method{} explicitly states that the LGN projects to cortical layer IV, but attributes the axons in that layer to the retina and selects (C) in 187 tokens. This error conflates the pathway's upstream information source with the origin of the direct projection.

\subsection{MathVista \#162: Recognizing but Not Applying a Lower Bound}

\begin{figure}[ht]
    \centering
    \includegraphics[width=0.99\linewidth]{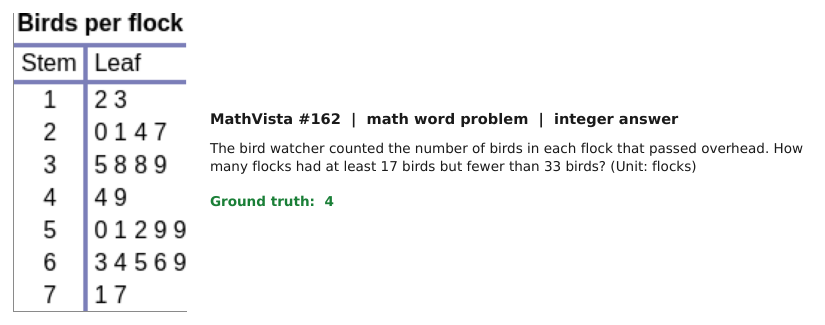}
    \caption{\textbf{MathVista~\#162: interval-constrained counting from a stem-and-leaf plot (reference answer: 4).} The qualifying values within $[17, 33)$ are $20, 21, 24, 27$. Base tests both boundaries and reports 4 in 1780 tokens. \method{} verifies only the upper bound ($<33$), incorrectly including $12$ and $13$ to answer 6 in 221 tokens. Transcripts are shown in Tables~\ref{tab:casef2base} and~\ref{tab:casef2copd}.}
    \label{fig:casef2}
\end{figure}

\begin{breakabletable}
    \caption{\textbf{Base} model response on MathVista~\#162 (cuts marked \textit{[\dots]}).}
    \label{tab:casef2base}
    \begin{tcolorbox}[
      breakable,
      fontupper=\footnotesize,
      title={Qwen3-VL-2B-Instruct (base) --- 1780 tokens, judged \textbf{correct}},
    ]
        To determine how many flocks had at least 17 birds but fewer than 33 birds, we need to analyze the given stem-and-leaf plot. \textit{[\dots]} The stem represents the tens digit, and the leaf represents the units digit.\\[2pt]
        \textbf{Stem 1}: Leaves are 2 and 3 $\rightarrow$ numbers: 12, 13. \textbf{Stem 2}: Leaves are 0, 1, 4, 7 $\rightarrow$ numbers: 20, 21, 24, 27. \textbf{Stem 3}: Leaves are 5, 8, 8, 9 $\rightarrow$ numbers: 35, 38, 38, 39. \textit{[\dots]}\\[2pt]
        Stem 1: 12, 13 $\rightarrow$ only 12 and 13 $\rightarrow$ both $<$ 17 $\rightarrow$ no. \textit{[\dots]} 20, 21, 24, 27 --- all of which are $\ge$ 17 and $<$ 33. \textit{[\dots]}\\[2pt]
        So there are 4 flocks with at least 17 birds but fewer than 33 birds. \textit{[\dots]} So the final answer is 4.
    \end{tcolorbox}
\end{breakabletable}

\begin{breakabletable}
    \caption{\textbf{\method{}} response on MathVista~\#162 (complete and unedited).}
    \label{tab:casef2copd}
    \begin{tcolorbox}[
      breakable,
      fontupper=\footnotesize,
      title={\method{} (ours) --- 221 tokens, judged \textbf{wrong}},
    ]
        To determine how many flocks had at least 17 birds but fewer than 33 birds, we analyze the stem-and-leaf plot.\\[2pt]
        \textbf{Stem 1} (10s). Leaves $2,3$ $\rightarrow$ Birds $12, 13$ (both $<33$). Count: 2 flocks.\\[2pt]
        \textbf{Stem 2} (20s). Leaves $0,1,4,7$ $\rightarrow$ Birds $20,21,24,27$ (all $<33$). Count: 4 flocks.\\[2pt]
        \textbf{Stem 3} (30s). Leaves $5,8,8,9$ $\rightarrow$ Birds $35,38,38,39$ ($\ge 33$). Not included.\\[2pt]
        Total flocks with 17--32 birds: $2 + 4 = 6$.\\[2pt]
        Answer: \textbf{6}.
    \end{tcolorbox}
\end{breakabletable}

The task requires counting entries within the half-open interval $[17,33)$ in a stem-and-leaf plot, where the qualifying values are $20,21,24,$ and $27$, giving a count of 4. Within Base's 1780-token response, the decisive step is applying both bounds: excluding $12$ and $13$ because they fall below 17 and retaining the four qualifying values. This step translates the conjunction of conditions in the question into an explicit filtering operation.

\method{} transcribes the data entries with high visual fidelity, but only enforces the upper threshold during filtering (recording entries as ``$<33$'' while omitting the lower bound ``$\ge 17$''). As a result, it includes $12$ and $13$ in the summation, yielding an inflated count of 6 in 221 tokens. Notably, \method{} cites ``at least 17'' in its opening and ``17--32'' in its summary, yet drops this condition during numerical filtering. The error lies in the gap between restating and applying the constraints: the response retains both bounds in its wording but enforces only the upper bound during counting.

\subsection{MMStar \#386: A Final Answer That Contradicts the Stated Direction}

\begin{figure}[ht]
    \centering
    \includegraphics[width=0.99\linewidth]{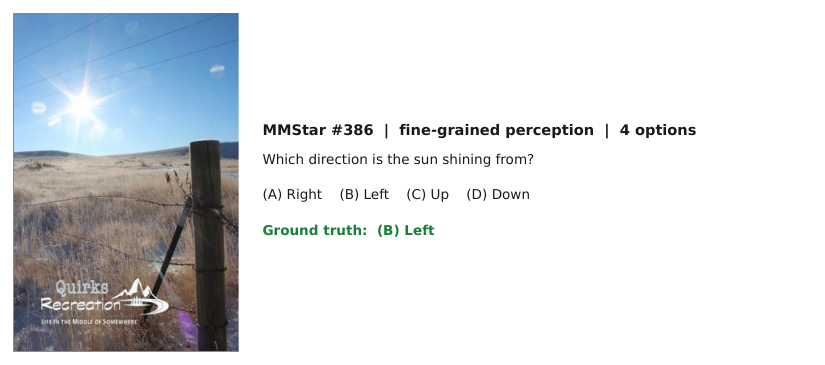}
    \caption{\textbf{MMStar~\#386: inferring illumination direction from the sun's visible position (reference answer: (B) Left).} The sun's starburst is visible in the upper-left portion of the image. Base identifies this position and selects (B) in 250 tokens. \method{} also states that the sun shines from the left, yet concludes with the opposite option (A) Right in a 95-token response. Model responses appear in Tables~\ref{tab:casef3base} and~\ref{tab:casef3copd}.}
    \label{fig:casef3}
\end{figure}

\begin{breakabletable}
    \caption{\textbf{Base} model response on MMStar~\#386 (cuts marked \textit{[\dots]}).}
    \label{tab:casef3base}
    \begin{tcolorbox}[
      breakable,
      fontupper=\footnotesize,
      title={Qwen3-VL-2B-Instruct (base) --- 250 tokens, judged \textbf{correct}},
    ]
        Based on the visual evidence in the image, we can determine the direction the sun is shining from.\\[2pt]
        The sun is depicted as a bright, intense light source in the upper left portion of the sky. The shadows cast by the fence post and the dry grass are angled towards the right side of the image.\\[2pt]
        \textbf{A. Right}: this would mean the sun is coming from the right, but the light is strongest on the left. \textbf{B. Left}: this is consistent with the sun being in the upper left. \textbf{C. Up}: this would create a high-angle light, not a strong horizontal glare like seen here. \textit{[\dots]}\\[2pt]
        The most logical conclusion is that the sun is shining from the \textbf{left}. Therefore, the correct answer is \textbf{B. Left}.
    \end{tcolorbox}
\end{breakabletable}

\begin{breakabletable}
    \caption{\textbf{\method{}} response on MMStar~\#386 (complete and unedited).}
    \label{tab:casef3copd}
    \begin{tcolorbox}[
      breakable,
      fontupper=\footnotesize,
      title={\method{} (ours) --- 95 tokens, judged \textbf{wrong}},
    ]
        Based on the image, the sun is shining from the \textbf{left}. Here's why:\\[2pt]
        --- The bright starburst is clearly visible in the upper-left portion of the sky.\\[2pt]
        --- The light is hitting the landscape from that direction, illuminating the left side of the field and casting shadows to the right.\\[2pt]
        --- The wires and fence posts are angled toward the right, which aligns with the sun's position.\\[2pt]
        Therefore, the correct answer is: \textbf{A. Right}.
    \end{tcolorbox}
\end{breakabletable}

The question asks for the direction of illumination in an outdoor scene. The sun is directly visible as a bright starburst in the upper-left portion of the image, providing a clear positional cue. Base identifies this location and correctly maps its judgment that the light comes from the left to option (B), completing the response in 250 tokens.

\method{} explicitly locates the sun in the upper left and states that the light comes from the left, yet ends with ``A. Right''. Its additional explanation based on the orientation of wires and fence posts provides no valid evidence for the illumination direction. The key error is that the final option reverses the direction stated earlier in the response.

\clearpage

\section{Limitations and Future Work}
\label{app:limitations}

\paragraph{Limitations.}
Empirical gains also exhibit task-dependent heterogeneity across multimodal domains. In the Qwen3-VL-8B$\to$2B distillation in Table~\ref{tab:main}, \method{} incurs modest trade-offs on benchmarks with high perceptual demands (losing 1.8 pp on HalluBench and 0.5 pp on RealWorldQA relative to Base). Besides, our empirical validation focuses on the Qwen3-VL and Qwen3.5 model families across dense and Mixture-of-Experts architectures. Systematically validating contrastive advantage across heterogeneous tokenizers, and non-English corpora remains an open empirical direction.

\paragraph{Future Work.}
Future investigations can advance this framework along two principal trajectories: (1) \textbf{Multi-Attribute Contrastive Conditioning}: extending paired teacher scoring beyond length to other orthogonal dimensions, such as safety alignment and linguistic style alignment; and (2) \textbf{Input-Adaptive Advantage Bounds}: designing instance-conditional clipping or complexity-aware budget schedulers that dynamically relax compression pressure when high visual ambiguity is detected.

\end{document}